\documentclass[nohyperref]{article}

\usepackage{microtype}
\usepackage{graphicx}
\usepackage{subfigure}
\usepackage{booktabs} 

\usepackage{hyperref}

\usepackage[accepted]{icml2022}

\usepackage{amsmath}
\usepackage{amssymb}
\usepackage{mathtools}
\usepackage{amsthm}

\usepackage[capitalize,noabbrev]{cleveref}

\theoremstyle{plain}

\theoremstyle{definition}

\theoremstyle{remark}

\icmltitlerunning{What Can Linear Interpolation of Neural Network Loss Landscapes Tell Us?}

\begin{document}

\twocolumn[
\icmltitle{What Can Linear Interpolation of Neural Network Loss Landscapes Tell Us?}

\begin{icmlauthorlist}
\icmlauthor{Tiffany Vlaar}{edi}
\icmlauthor{Jonathan Frankle}{mit}
\end{icmlauthorlist}

\icmlaffiliation{edi}{Department of Mathematics, University of Edinburgh, Edinburgh, United Kingdom}
\icmlaffiliation{mit}{CSAIL, MIT, Cambridge, United States}

\icmlcorrespondingauthor{Tiffany Vlaar}{Tiffany.Vlaar@ed.ac.uk}
\icmlcorrespondingauthor{Jonathan Frankle}{jfrankle@csail.mit.edu}

\icmlkeywords{} 

\vskip 0.3in
]

\printAffiliationsAndNotice{} 

\begin{abstract}
Studying neural network loss landscapes provides insights into the nature of the underlying optimization problems.
Unfortunately, loss landscapes are notoriously difficult to visualize in a human-comprehensible fashion.
One common way to address this problem is to plot linear slices of the landscape, for example from the initial state of the network to the final state after optimization.
On the basis of this analysis, prior work has drawn broader conclusions about the difficulty of the optimization problem.
In this paper, we put inferences of this kind to the test, systematically evaluating how linear interpolation and final performance vary when altering the data, choice of initialization, and other optimizer and architecture design choices.
Further, we use linear interpolation to study the role played by individual layers and substructures of the network.
We find that certain layers are more sensitive to the choice of initialization, but that the shape of the linear path is not indicative of the changes in test accuracy of the model.
Our results cast doubt on the broader intuition that the presence or absence of barriers when interpolating necessarily relates to the success of optimization.
\end{abstract}

\section{Introduction}

Neural network loss landscapes are difficult to visualize due to their high-dimensionality and the complicated nature of the actual optimization path.
This motivated the use of the loss along the linear path between the initial and final parameters of a neural network as a crude yet simple measure of the loss landscape \cite{Goodfellow}.  In this work we  revisit this 1D linear interpolation technique and address whether the shape of the path reflects the test accuracy of the model and can inform training.

\textbf{Linear interpolation from beginning to end.}
\citet{Goodfellow} observed that for fully-connected and convolutional networks with maxout \cite{maxout} trained on MNIST data the loss decays monotonically along the linear path between their initial and final state.
The absence of obstacles along the linear path led them to conclude that ``these tasks are relatively easy to optimize.''
This result has been cited widely as an indication of the ease of training (e.g., \citet{Li2018intrinsic,McCandlish2018,Fort2019}) and the linear interpolation technique itself was used in many papers \cite{Huang2017,Keskar2017,SWA,Jastrzebski2018,Hao2019}. In this paper we address empirically how meaningful the use of this linear path actually is.
The exact definition of training tasks being ``easy to optimize'' is an open question and optimization choices directly influence which linear path we observe. It is also arguable whether we actually want tasks to be easier to optimize; lowering the amount of training data or reducing regularization simplifies training but lowers test accuracy.

The work by \citet{Goodfellow} was recently revisited and extended by \citet{Frankle2020} and \citet{Lucas2021} for a range of modern neural network models, such as ResNet and VGG architectures, on image data sets.
\citet{Frankle2020} observed for default parameter settings that the loss often remained at the level of random chance until close to the optimum 
for these models, different than the behavior observed by \citet{Goodfellow}.
In addition to concurrently confirming this result, \citet{Lucas2021} found that the monotonic decay property observed by \citet{Goodfellow} was often maintained when BatchNorm was removed and non-adaptive optimizers were used.
On this basis, they hypothesized that ``large distances moved in weight space encourage non-monotonic interpolation.''

In our work, we interrogate conjectures stated in prior work that the shape of loss along the linear path relates to the ``success'' of optimization (which we measure in terms of test accuracy) or other aspects of optimization (e.g., distance travelled).
We systematically study the influence of various optimizer and architecture design choices on the shape of the linear path and the test accuracy of the final model, and examine interpolation for individual layers in addition to the entire network.
An overview of our results is shown in Table \ref{table:overview}.
Our main finding is that there are situations that both support and violate the aforementioned intuitions on the shape of the linear path.
As such, we recommend caution when using this analysis to infer other information about the nature of the optimization problem.

\begin{table*}
\begin{center}
\caption{The effect of various interventions in training in (1) the shape of loss over the linear path between the initial and final state of the network and (2) the final test accuracy of the network.
Different shapes of loss over the linear path are: no barrier (\textcolor{blue}{\textit{NB}}), barrier (\textcolor{red}{\textit{B}}), and plateau (P).
If the height of layer-wise barriers changes, we denote this as \textcolor{red}{\textit{H-LB}} (higher) and \textcolor{blue}{\textit{L-LB}} (lower layer-wise barriers). \\ 
}\label{table:overview}
{
\resizebox{0.8\textwidth}{!}{
\begin{tabular}{c|c|c|c}
   Category & Intervention & Shape of the linear path & Test accuracy \\ \hline 
    & Pre-train full model (Fig. \ref{fig:pretrainedRL}A) & \textcolor{blue}{NB} & \textcolor{blue}{Better} \\ 
 \textbf{Initialization} & Pre-train on random labels (Fig. \ref{fig:pretrainedRL}B) & \textcolor{red}{H-LB} & \textcolor{red}{Worse} (if no weight decay)  \\
  \textit{(Sec. \ref{sec:initialization})} & Height of the barrier initialization (Fig. \ref{fig:heightofbarrier_withoutwd}) & \textcolor{blue}{NB} & \textcolor{blue}{Better} \\
& Partial (pre-)training (Fig. \ref{fig:pretrainsomelayers}) & \textcolor{red}{B}/\textcolor{blue}{NB}/P & \textcolor{red}{Often worse} \\ 
& Partial random label pre-training (Fig. \ref{fig:advinitsomelayers}) & \textcolor{red}{B}/\textcolor{blue}{NB} & \textcolor{red}{Often Worse} 
\\ \hline 
   \textbf{Data} & Less data (Fig. \ref{fig:numberoftrainingdatadataaug}) & \textcolor{blue}{NB} / \textcolor{blue}{L-LB} & \textcolor{red}{Worse}\\
      \textit{(Sec. \ref{sec:data})} & No data augmentation (Fig. \ref{fig:numberoftrainingdatadataaug}) & \textcolor{blue}{L-LB} & \textcolor{red}{Worse}\\ \hline
 & Less weight decay (Fig. \ref{fig:WD}) &  \textcolor{blue}{NB} / \textcolor{red}{B} & \textcolor{red}{Worse} \\ 
  & More weight decay  (Fig. \ref{fig:WD}) &  P & \textcolor{red}{Worse} \\ 
\textbf{Optimizer} & Fixed learning rate $h$ = 0.01 (Fig. \ref{fig:fixedlr}) & P & \textcolor{blue}{Better} \\ 
 \textit{(Sec. \ref{sec:optimizer})} & Fixed learning rate $h\neq$ 0.01  (Fig. \ref{fig:fixedlr}) & \textcolor{blue}{NB} & \textcolor{red}{Worse} \\ 
 & Smaller initial $h$ for conv blocks with barriers (Fig. \ref{fig:linearintp_difflr}) & \textcolor{blue}{NB} &  \textcolor{red}{Worse} \\\hline
& Depth (Fig. \ref{fig:compareresnets}) & \textcolor{red}{H-LB} & \textcolor{blue}{Better} \\ 
   \textbf{Model} & No batch normalization (Appx. \ref{sec:appxmodel}) & \textcolor{blue}{NB} & \textcolor{red}{Worse}   \\ 
\textit{(Sec. \ref{sec:model})} & MLPs: overparameterize (Appx. \ref{sec:appxmodel}) & \textcolor{blue}{NB} & \textcolor{blue}{Often better} \\ 
\end{tabular}
}
}
\end{center}

\vspace{-0.37cm} 
\end{table*}

\textbf{Our findings:}
\begin{itemize} 
    \item Pre-training on ImageNet consistently removes the presence of barriers for ResNet architectures trained on CIFAR-10 data, whereas adversarial initialization on random labels increases barriers. The former typically increases the final test accuracy, whereas (in the absence of weight decay) the adversarial initialization lowers it.
    \item Layers have different levels of sensitivity to the choice of initialization. We introduce the concept of partial pre-training, where we set some layers to a trained (on CIFAR-10) or pre-trained (on ImageNet) state, while using random initialization for others. We use this setting as initialization and train on CIFAR-10. We find that partial pre-training generally leads to worsened test accuracy and (uncorrelated) affects the shape of the linear path. 
    \item 
    The amount of weight decay used during training directly influences both the shape of the linear interpolation path
    and the final test accuracy, but there is no correlation between them. 
    \item The distance between the initial and final parameter state is not a reliable indicator of non-monotonic behaviour along the linear path.
 \item The shape of the linear path from initial to final parameter state is not a reliable indicator of test accuracy.
\end{itemize}

\section{Related Work}

\textbf{Interpolation on the loss landscape.} 
Interpolation between networks in various forms has been a valuable tool for gaining insight into the structure of the optimization landscape.
\citet{Frankle2020} 
found that although the loss did not increase 
from initial to final state, barriers did appear along the linear path from later iterations to the final state. Other work focused on interpolations between different optima: \citet{Draxler2018} and \citet{Garipov2018} showed that there exist non-linear paths with (nearly) constant low-loss that connect a pair of minima trained from different random initializations.
\citet{FortJastrzebski2019} generalized this to show that there exist $m$ low-loss connectors between $(m+1)$-tuples of optima, again using non-linear paths.
\citet{Neyshabur2020} used linear interpolation to study transfer learning. 
They did not observe barriers along the linear path between the final states of two ResNet-50 models both trained from pre-trained, while barriers did occur between two models trained from scratch (even when trained with the same random initialization). 
They interpreted this as ``pre-trained weights guide the optimization to a flat basin of the loss landscape.''
We comment on the role of pre-training
in Section \ref{sec:initialization}.

\textbf{Loss landscape visualization.} To improve upon linear interpolation, 2D or 3D visualizations of the loss landscape were made by \citet{Goodfellow,Im2016,Li2018,Hao2019}. The reduction of the high-dimensional loss landscape into 2D or 3D slices requires a choice of directions. The chosen directions strongly affect the resulting observed behaviour. Although the linear interpolation method also sacrifices information by taking a 1D slice, it does so with perfect fidelity by considering the path between initial and final state. Linear interpolation is seen as ``a simple and lightweight method to probe neural network loss landscapes'' \cite{Lucas2021} and therefore remains frequently used, e.g. by \citet{Keskar2017,Jastrzebski2018,Lucas2020,Neyshabur2020}. 

\textbf{Role of layers.}
\citet{Zhang2019} studied the role of different layers 
by training a neural network and then re-setting specific layers to their initial value or a random value, while keeping the other layers fixed at their final state. 
They observed that certain critical layers are much more sensitive to this perturbation. \citet{Chatterji2020} extended this analysis by studying linear interpolation for specific modules.
They relate their concept of ``module criticality'' with high robustness to noise and valley width.
\citet{Neyshabur2020} 
studied both the direct and optimization path from initial to final state for modules of pre-trained models and found that later layers have tighter valleys.

\textbf{Our novel contributions.} Previous work studied the shape of the linear path \cite{Frankle2020,Lucas2021}, but did not interrogate the connection with the success of optimization. Dating back to \citet{Goodfellow} intimating that an absence of barriers along the linear path means that ``tasks are relatively easy to optimize", numerous works have implicitly relied on the presence of such a connection to make other claims \cite{McCandlish2018,Li2018,Fort2019,Hao2019,Lucas2020, Neyshabur2020}. In our work, we study exhaustively if such a connection exists by systematically altering initialization, data, and other optimizer and architecture design choices. We also study the hypothesis by \citet{Lucas2021} that ``large distances moved in weight space encourage non-monotonic interpolation''.
Further, we introduce several novel modes of analysis, such as initializing from the height of the barrier and using linear interpolation to study the role played by individual layers and substructures of the network. In particular,  
we illustrate the sensitivity of different layers to the choice of initialization and demonstrate the adversarial effect of partial pre-training (Section \ref{sec:initialization}).

\section{Linear Interpolation from Start to Finish}
We use the 1D linear interpolation technique \cite{Goodfellow} to study the linear path between the initial and final state of the model. We introduce this technique and how we study the linear path layer-wise in Section \ref{sec:methodology}. In Section \ref{sec:barriers} we discuss which different shapes of this linear path we observe and compare our results with 
the literature \cite{Goodfellow,Frankle2020,Lucas2021}.

\subsection{Methodology}\label{sec:methodology} 

\textbf{Training.}
We focus on a ResNet-18 \cite{resnet} architecture with batch normalization trained for 100 epochs on CIFAR-10 data \cite{cifar10} using SGD with momentum ($0.9$) and weight decay (5e-4) using PyTorch \cite{Pytorch}.
We use initial learning rate $h = 0.1$ that drops by 10x at epochs 33 and 66.
For pre-trained settings, we use initial learning rate $h = 0.001$ that drops by 10x after 30 epochs. Results are averaged over 10 runs unless indicated otherwise.
By modifying different aspects of this training problem, we will study the role of the initialization, data, optimizer, and model throughout this paper.
We also consider other architectures in the main body and supplement, including other ResNets, VGG architectures, and multi-layer perceptrons. 

\textbf{Linear interpolation measure for the full model.} 
Consider a $L$-layer neural network with parameters $\theta = (\theta^{(0)},\dots,\theta^{(L)})$. We use $\theta_i$ to refer to the initial state of these parameters and $\theta_f$ to refer to the parameters after training using algorithm $\mathcal{T}_{t}(\theta_i,\mathcal{D})$ on dataset $\mathcal{D}$ for $t$ steps. Following \citet{Goodfellow}, a linear interpolation path between $\theta_i$ and $\theta_f$ is created as follows:
\begin{align}
\theta_{\alpha} = (1-\alpha)\theta_i+\alpha\theta_f~~\textrm{for}~~\alpha\in [0,1].
\end{align}
To examine the loss landscape along this path, we plot the loss for a discrete set of values of $\alpha$ from 0 (initial state) to 1 (final state).
One can extend this technique to evaluate the linear path between the state of the model at different steps in training, where $\theta_i$ and $\theta_f$ are replaced by the model states at the considered steps. 
One can also study the path between $\theta_f$ and a different random initialization $\theta'_i$, which is separately sampled from the initialization distribution.

\textbf{Layer-wise linear interpolation.} 
We also study linear interpolation in a layer-wise fashion: we vary a single layer (or convolutional block) from initial to final state while keeping all other parameters fixed at their final state.
Concretely, for an $L$-layer network where we vary layer $\ell$:
\begin{align}
     \theta^{(\ell)}_{\alpha} = (1-\alpha)\theta^{(\ell)}_0+\alpha\theta^{(\ell)}_f, \ \ \ \theta^{(k)}_{\alpha} = \theta^{(k)}_f, \ k \neq \ell.
\end{align}

This technique was first proposed by \citet{Chatterji2020}, who found that certain ResNet layers were more robust to parameter perturbations.
In this work, we use the layer-wise linear path to study the role played by substructures of the network. 
We will vary convolutional blocks as a whole. 

\subsection{Appearance of Barriers in the Loss Landscape}\label{sec:barriers}

\begin{figure*}[t]
    \centering
    \includegraphics[width=0.43\linewidth]{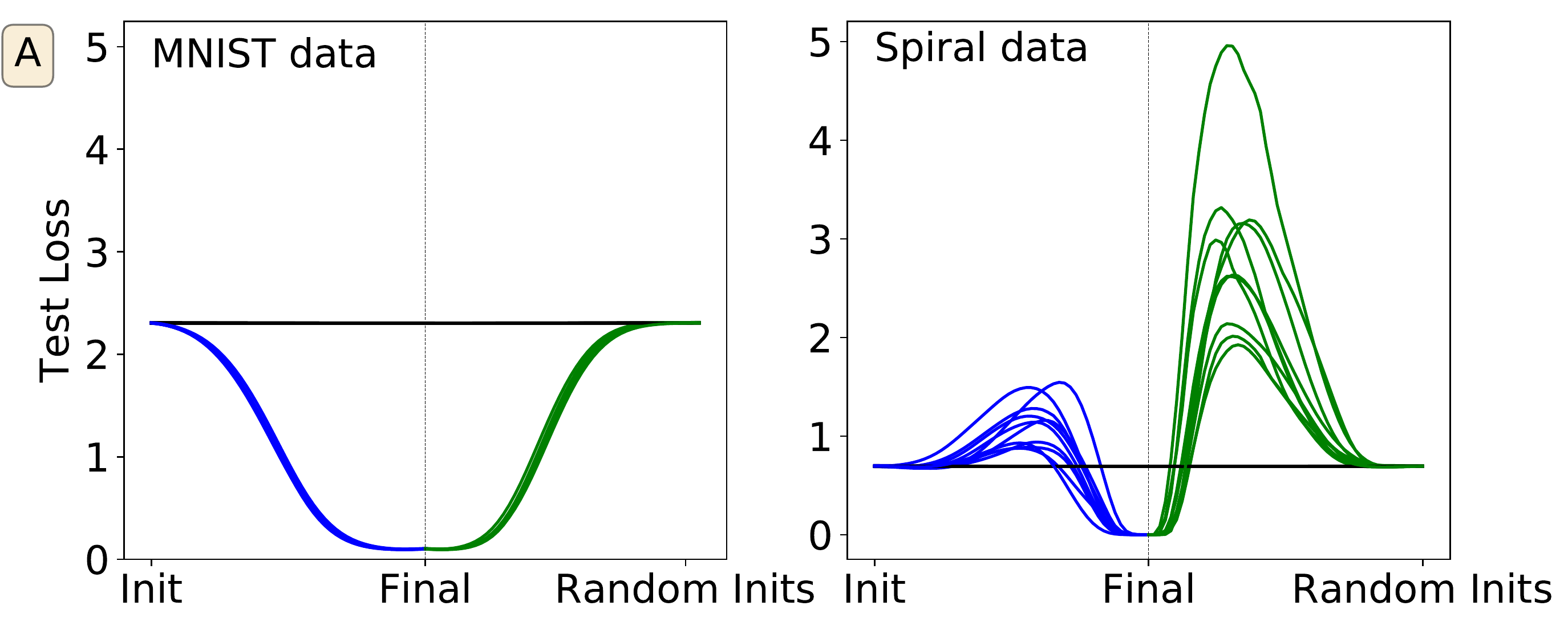} \ \ \  
     \includegraphics[width=0.25\linewidth]{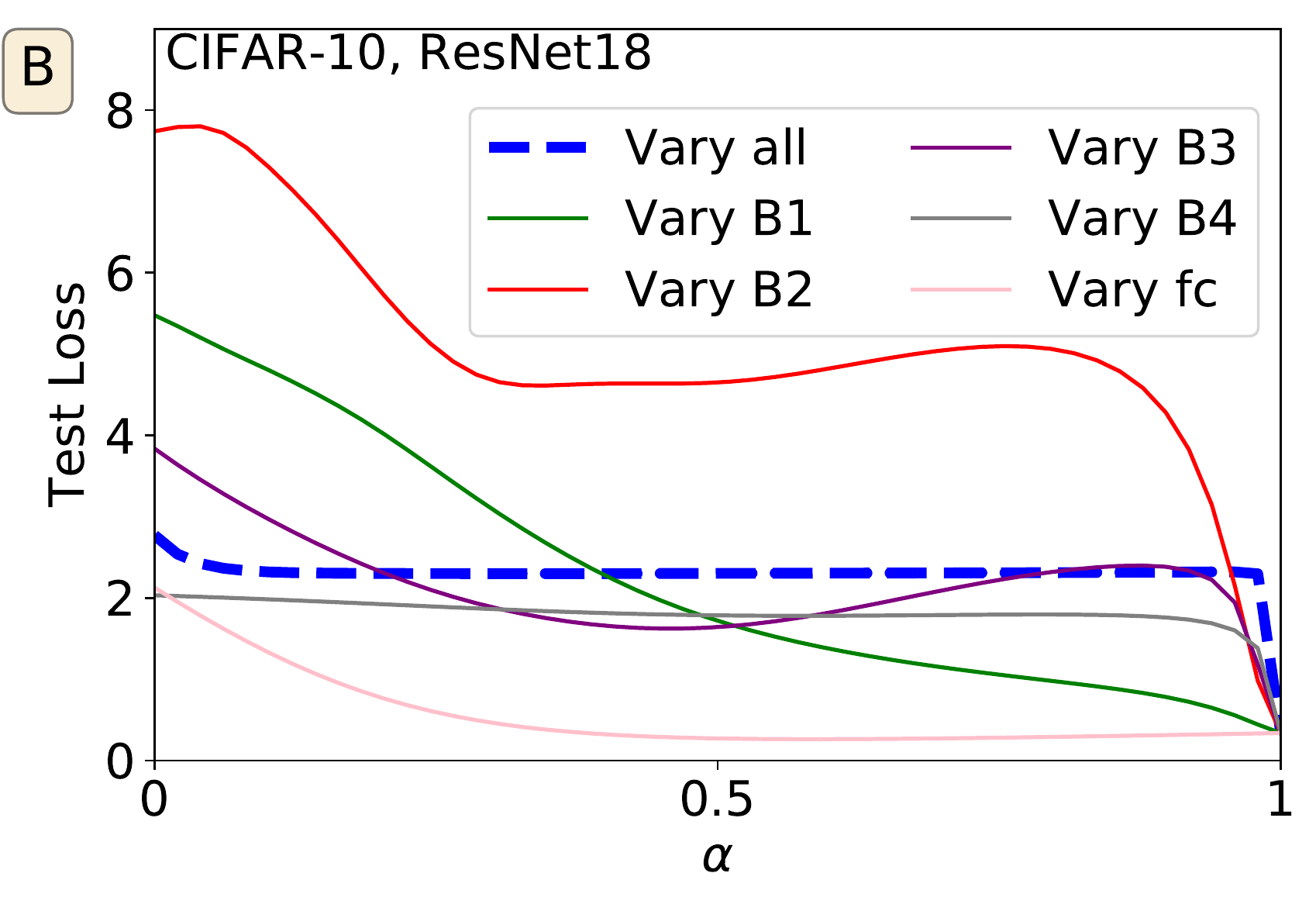}
   \vspace*{-4mm}
    \caption{\textit{(A)} Left: MNIST data. Right: spiral data. Test loss over 10 runs for a two hidden layer MLP along the linear path between: 
    $\theta_i$ and 
    $\theta_f$ (blue), $\theta_f$ and 
    $\theta'_i$ (green), $\theta_i$ and $\theta'_i$ (black).
    \textit{(B)} Test loss when interpolating for convolutional blocks (denoted as B), fully connected layer (fc), or the entire network (blue dashed line) for a ResNet-18 architecture on CIFAR-10 data, averaged over 10 runs.
}
    \label{fig:mnistspiralresnet}
\end{figure*}

\begin{figure*}
    \centering
    \includegraphics[width=0.8\linewidth]{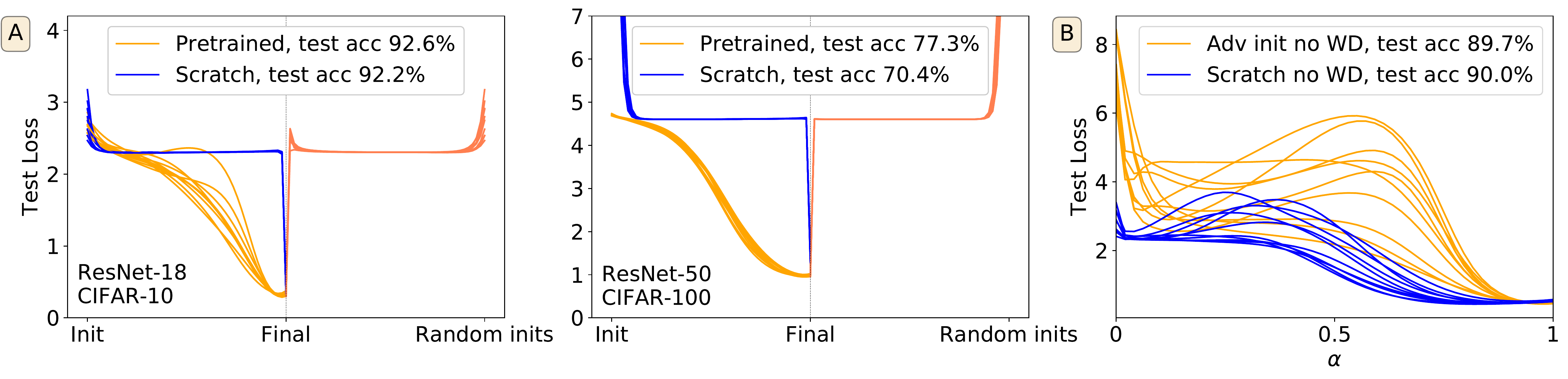}
    \vspace{-4mm}
    \caption{\textit{(A)}  Test loss between $\theta_i$ and $\theta_f$ 
    of a ResNet-18 for CIFAR-10 (left) and ResNet-50 for CIFAR-100 (middle), when trained from scratch (blue) or pre-trained on ImageNet (orange) over 10 runs. Orange-red lines represent the linear path between $\theta_f$ (trained from pre-trained) to $\theta'_i$. 
    \textit{(B)}  Test loss between $\theta_i$ and $\theta_f$ 
    of a ResNet-18 trained with SGD without weight decay (WD) on CIFAR-10 either from scratch (blue) or from a network pre-trained on 100\% random labels (orange).
    }
    \label{fig:pretrainedRL}
\end{figure*}

\textbf{Defining barriers.}
We consider the linear path to contain a \textit{barrier} if it exhibits a monotonic increase in loss.
\citet{Goodfellow} observed that, for fully-connected networks on MNIST data, loss decays monotonically along the linear path between initial and trained model, i.e., there are no barriers.
Figure \ref{fig:mnistspiralresnet}A (left) reproduces this behaviour for an MLP with two hidden layers trained on MNIST (blue) (full details in Appendix \ref{sec:implementation}); moreover, the same monotonic decrease occurs when interpolating between these final weights and \emph{any} random initialization (green).
However, this barrier-free linear interpolation is not a universal phenomenon.
As a counterexample, when using the exact same architecture and optimizer but a different dataset (the spiral dataset), barriers do appear along the linear path (Figure \ref{fig:mnistspiralresnet}A, right), in fact rising above the level of loss at initialization; barriers rise even higher when interpolating to other initializations.

\textbf{Barriers in modern neural network architectures.}
\citet{Frankle2020} and \citet{Lucas2021} updated the results of \citet{Goodfellow} by studying linear interpolation in modern vision settings.
Both observed that, in many cases, ``loss plateaus and error remains at the level of random chance...until near the optimum'' \cite{Frankle2020}; that is, loss remains flat, neither monotonically decreasing or encountering barriers.
Our results for a ResNet-18 on CIFAR-10 agree with these findings when interpolating for the entire network as the dashed blue line in Figure \ref{fig:mnistspiralresnet}B illustrates.

\textbf{Block-wise interpolation.}
Individual convolutional blocks behave differently from the full network.
As the solid lines in Figure \ref{fig:mnistspiralresnet}B show, different blocks take on a variety of behaviors including barriers and monotonic decay.
This suggests that it may be valuable to study the connection between linear interpolation and other properties of the network at the finer granularity of individual structural components rather than at the coarse granularity of the entire network.

\section{The Role of Initialization}\label{sec:initialization} 
The choice of initialization affects the path that the model follows and the optimum it finds.
It is therefore natural to believe that it also affects the nature of the loss when linearly interpolating from start to finish, a relationship we study in this section.
We observe an intuitive relationship between initialization and barriers: actions that make the task easier (e.g., pre-training the model or initializing at the barrier) remove barriers, while those that make the task harder (e.g., initializing adversarially) increase the size of barriers.
Pre-training only certain layers can both create barriers and worsen test accuracy, although not in a correlated fashion.

\textbf{Pre-training.}
Training a model from a pre-trained state (on ImageNet) causes the loss along the linear path to monotonically decay 
and 
improves test accuracy (Figure \ref{fig:pretrainedRL}A).
This is distinct from the behavior when interpolating to a new random initialization, which exhibits a plateau. This result aligns with the intuition 
that pre-training simplifies optimization \cite{Hao2019,Neyshabur2020}.

\textbf{Adversarial initialization.} 
We perform adversarial initialization by training on 100\% random labels until 100\% training accuracy is reached and use this state as initialization for training. In the absence of weight decay,  pre-training a ResNet-18 on random labels lowers the final test accuracy \cite{badglobalminima} and 
increases the barrier height between initial and final state (Figure \ref{fig:pretrainedRL}B). 
This directly opposes the effect of pre-training (Figure \ref{fig:pretrainedRL}A), suggesting adversarial initialization complicates optimization.

\begin{figure*}[t]
     \centering
    \includegraphics[width=0.6\linewidth]{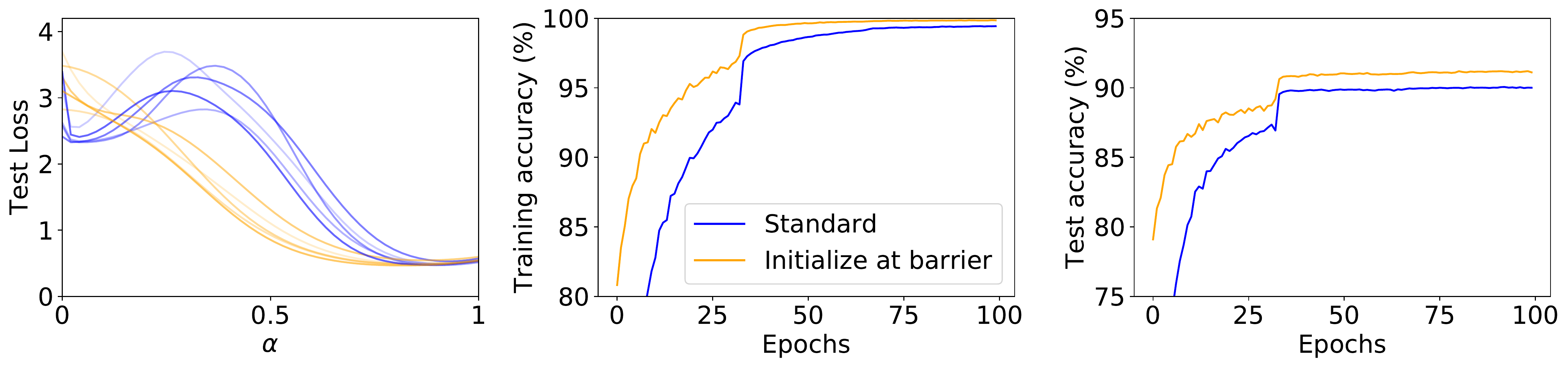}
    \vspace{-4mm}
         \caption{ResNet-18 on CIFAR-10
         without weight decay. Left: test loss between $\theta_i$ and $\theta_f$, when 
         trained either from scratch (blue) 
     or from a initialization corresponding to the height of the barrier along each blue line (orange). Initialization at the height of the barrier removes the presence of a barrier (left), speeds up training (middle), and improves the test accuracy (right), averaged over 5 runs. }
     \vspace{-3mm}
   \label{fig:heightofbarrier_withoutwd}
 \end{figure*}

\begin{figure*} 
    \centering
    {\small
      \begin{tabular}[b]{c|c}
      Method & Test acc (\%)  \\ \hline
   T-All but RI-1      & 91.79 $\pm$0.23\\
   T-All but RI-2      & 91.83 $\pm$0.21\\
   T-All but RI-3      & 92.35 $\pm$0.20\\
   T-All but RI-4      & 90.97 $\pm$0.31\\ \hline
   P-All but RI-1      & 89.97 $\pm$0.13\\
   P-All but RI-2      & 89.91 $\pm$0.21\\
   P-All but RI-3      & 91.78 $\pm$0.22\\
   P-All but RI-4      & 92.78 $\pm$0.22\\
    \end{tabular}}%
    \ \ \ 
    \includegraphics[width=0.5\linewidth]{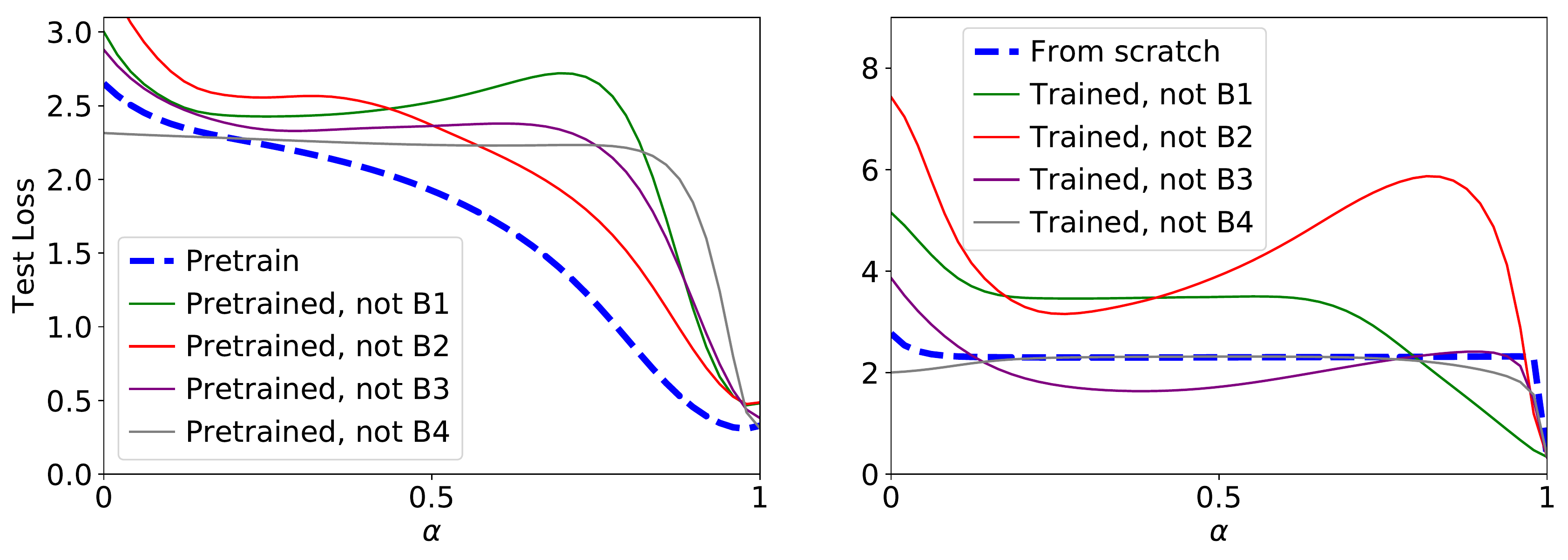}
        \vspace{-3mm}
    \caption{We set a ResNet-18 to a pre-trained \textit{(P)} (middle) or trained \textit{(T)} (right) state, but re-set a specific convolutional block \textit{(B)} to a random initialization \textit{(RI)}. Then we train as usual on CIFAR-10. This often affects the test accuracy of the final model (left). Sometimes barriers appear along the linear path for the full net between $\theta_i$ and
    $\theta_f$ (middle/right), averaged over 10 runs.}
    \label{fig:pretrainsomelayers}
    \vspace{-3mm}
\end{figure*}

\textbf{Height of the barrier initialization.}
The linear path of a ResNet-18 trained without weight decay on CIFAR-10 exhibits barriers for some random seeds. 
For the runs that exhibit barriers, we save the model state at the height of the barrier and use this as initialization for training to study if there exists a barrier between this 
state and the new optimum. 
Although it was initialized at a higher loss than occurs at random initialization, the resulting network obtains a higher test accuracy than the network trained from scratch and its linear path is barrier free (Figure \ref{fig:heightofbarrier_withoutwd}).\footnote{
Using other states along the linear path, which are 
far enough from the original initialization, as initialization often delivers similar improvements on the final test accuracy (see Appendix \ref{sec:appxinit}).} This mirrors the pre-training results (Figure \ref{fig:pretrainedRL}A), suggesting initialization at the height of the barrier aids optimization. 

\textbf{Partial pre-training.} 
\citet{Zhang2019} measured the change in test accuracy when re-setting specific layers of a trained neural network to their initial state and found large differences across layers.
To extend this work, one can study the loss from the initial state of a specific layer to its final state, while keeping all other layers fixed at their final state \cite{Chatterji2020}. We found that the shape of this path greatly varies per convolutional block of a ResNet-18 (Figure \ref{fig:mnistspiralresnet}B). But what these studies do not address is the effect of different layer-wise (or block-wise) initialization on training itself.
We thus introduce the concept of \textit{``partial (pre-)training"}: we first train on CIFAR-10 (or pre-train on ImageNet) and then re-set a specific convolutional block to its initial (random) state, while keeping the other parameters at their (pre-)trained state.
We then use this state of the model as initialization and train as usual on CIFAR-10 data. 

We find that while pre-training a ResNet-18 leads to higher test accuracy (Figure \ref{fig:pretrainedRL}A), partial pre-training 
often leads to a lower accuracy of the final trained network (Figure \ref{fig:pretrainsomelayers}, left) compared to training the net from scratch. Further, while pre-training the full net leads to monotonic decay along the linear path between initial and final state (Figure \ref{fig:pretrainedRL}A), partial pre-training often generates barriers (Figure \ref{fig:pretrainsomelayers}). 
The sensitivity of the network to partial pre-training varies per convolutional block and also between the trained and pre-trained setting (e.g., when re-setting convolutional block 4 (RI-4), using a trained state for the rest of the net (T-All) strongly affects test accuracy, but using a pre-trained state (P-All) does not). We conclude that the choice of initialization for different convolutional blocks strongly affects test accuracy after training and changes the nature of optimization. It is remarkable that using a random initialization for the full net leads to higher test accuracy than using a partially pre-trained or partially trained net as initialization. 

\begin{figure}[t]
    \centering
    \includegraphics[trim={0 0cm 25cm 0.8cm},clip,width=0.45\linewidth]
    {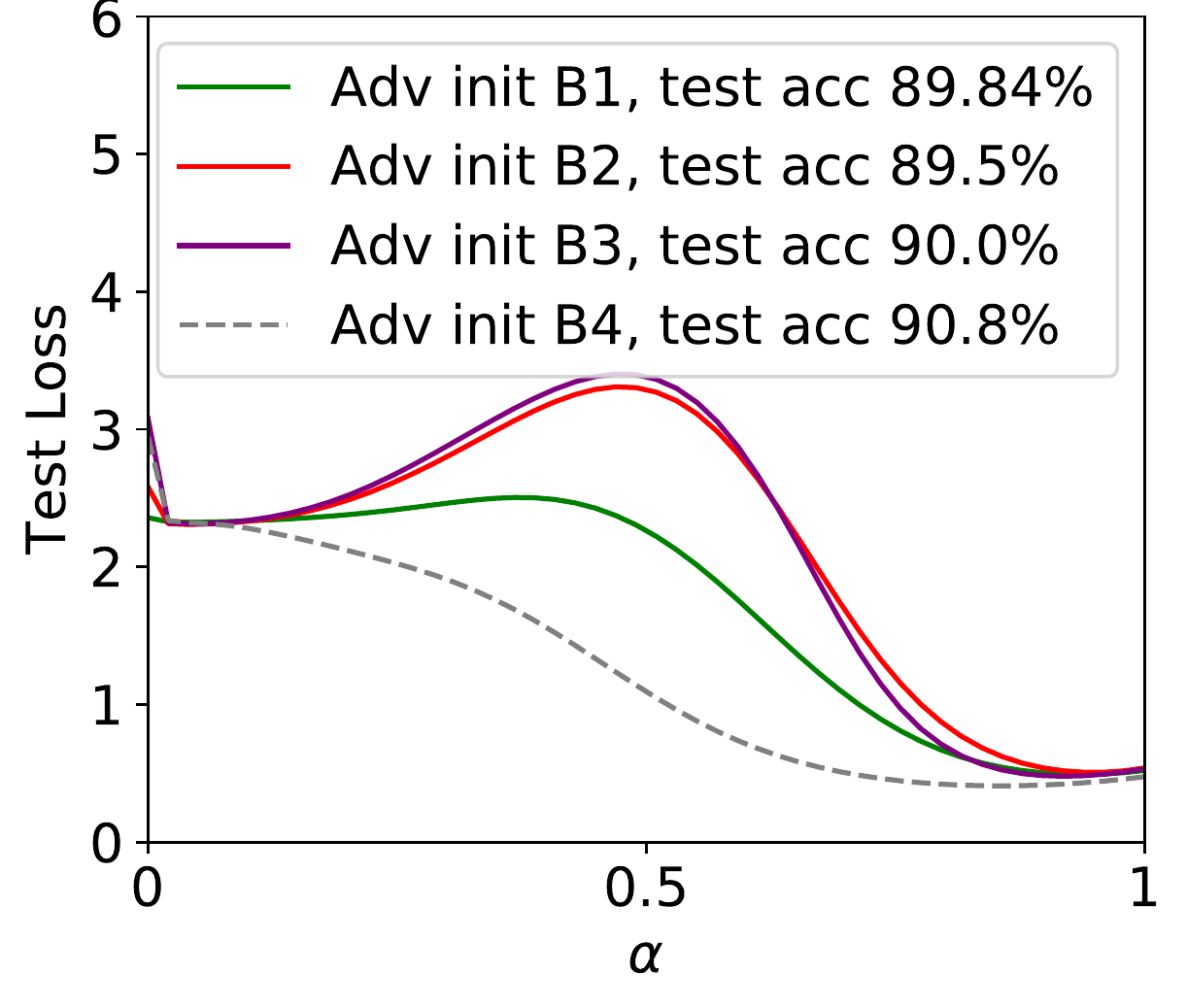}
   \vspace{-6mm}
    \caption{ResNet-18 on CIFAR-10 without weight decay. Test loss between $\theta_i$ and $\theta_f$ 
    when one conv block (B) was adversarially initialized using random labels, 
   averaged over 10 runs.} 
    \label{fig:advinitsomelayers} \vspace*{-0.4cm}
\end{figure}

\begin{figure*}
\centering
    \includegraphics[width=0.72\linewidth]{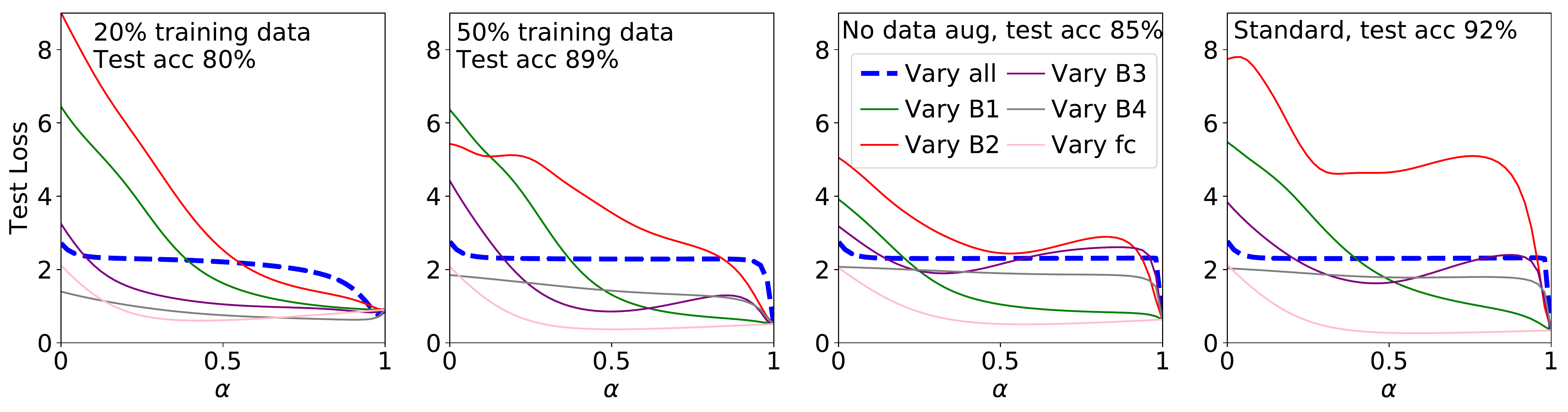}
    \vspace{-4.5mm}
    \caption{Test loss between $\theta_i$ and $\theta_f$ for ResNet-18 on CIFAR-10 when varying the amount of training data and whether data augmentation (horizontal flips, random crops) is used. Averaged over 10 runs.
    }
   \vspace{-4mm}
    \label{fig:numberoftrainingdatadataaug} 
\end{figure*}

\textbf{Block-wise adversarial initialization.} 
To further explore the effect of using different initializations for individual convolutional blocks on the linear path and test accuracy, we set one convolutional block to a random label adversarial initialization while using a standard random initialization for the rest of the net. We then re-train using this initialization. 
We find that while adversarial initialization for the first convolutional blocks lowers the test accuracy, using adversarial initialization for convolutional block 4 slightly increases the final test accuracy compared to training from scratch (Figure \ref{fig:advinitsomelayers}).
This is also reflected by the shape of the linear path: when using an adversarial initialization for convolutional block 4 the linear path exhibits monotonic decay while others settings exhibit barriers.\footnote{Similar results are obtained for different levels of weight decay (see Appendix \ref{sec:appxinit}).}

\textbf{Summary.} 
Whereas the use of pre-training removes the presence of barriers and increases test accuracy, the use of adversarial initialization increases barriers and lowers test accuracy. Initializing neural networks at the 
height of the barrier along their initial to converged state 
leads to monotonic decay along the linear path, speeds up training, 
and improves test accuracy. 
Further, we find that certain convolutional blocks are more sensitive to partial pre-training than others, but the change in test accuracy is not correlated with the shape of the linear path.
Whereas pre-training of the full net increases test accuracy and leads to monotonic decay, partial pre-training typically worsens the test accuracy of the resulting net and can generate barriers. 

\section{The Role of the Dataset}\label{sec:data}
As illustrated in Figure \ref{fig:mnistspiralresnet}A, where we used the same architecture, initialization, and optimizer to train on two different datasets, the data has a direct influence on the behaviour when linearly interpolating.
In this section, we further investigate the effect of data on interpolation.

\textbf{Number of examples.}
When only a subset of the training data set is used throughout training, we expect optimization to be easier and faster, yet the test accuracy of the final model is typically lowered.
We find that using a smaller amount of CIFAR-10 training data induces monotonic decay when interpolating 
and lowers test accuracy (Figure \ref{fig:numberoftrainingdatadataaug}, left). 

\textbf{Role of data augmentation.} Throughout this work we use a horizontal flip and random crop as data augmentation for CIFAR-10.
Removing data augmentation lowers the layer-wise barriers along the linear path (Figure \ref{fig:numberoftrainingdatadataaug}, center right).
In Appendix \ref{sec:appxdata}, we show that increasing data augmentation increases the height of layer-wise barriers, while removing data augmentation when using only a subset of the training data 
further reduces the presence of layer-wise barriers.

\textbf{Summary.}
The shape of the layer-wise linear path is affected by changes to the data, but the shape of the full model linear path is less representative (blue dashed lines, Figure \ref{fig:numberoftrainingdatadataaug}).
Reducing the complexity of the task (e.g., reducing the amount of data or removing augmentation) lowers or removes layer-wise barriers.
The setting with the highest layer-wise barriers reaches the highest test accuracy, and the setting without barriers reaches the lowest test accuracy.

\begin{figure*}[t]
\begin{minipage}{0.55\textwidth}
    \centering
    \includegraphics[width=\linewidth]{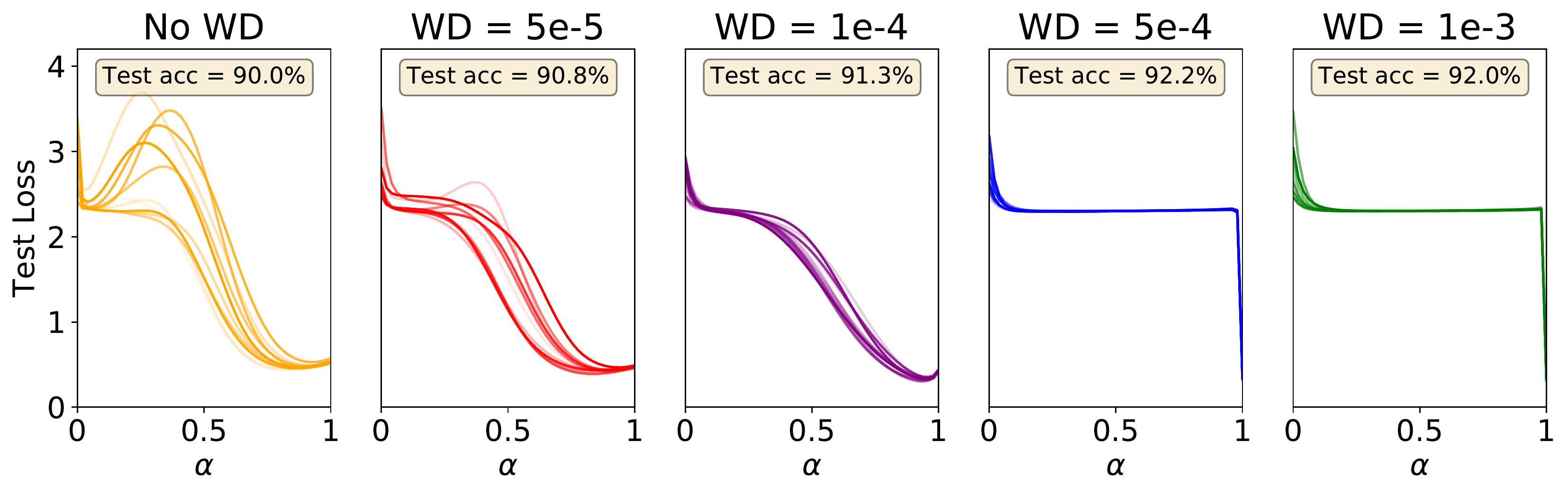} 
    \vspace{-3.5mm}
  \ \ \ \ \ \ \ \ \ \   \includegraphics[clip, trim=0cm 6cm 0cm 6cm,
    width=0.4\textwidth]{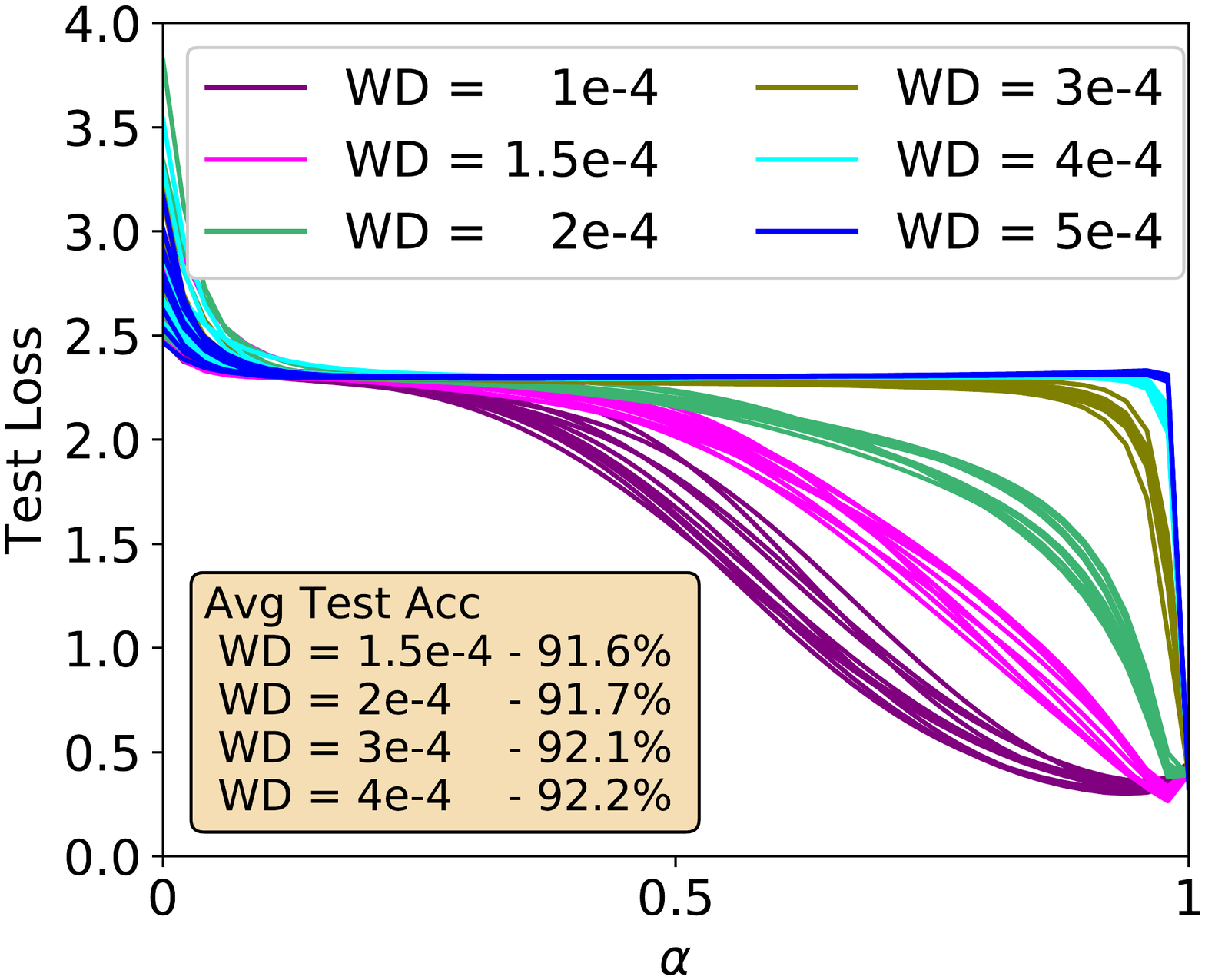}%
    \vspace{-2.7mm}
    \caption{Effect of the amount of weight decay (WD) used for training ResNet-18 on CIFAR-10 on the linear path and test accuracy. The highest test accuracy is reached using WD =5e-4 (blue).
    }
    \label{fig:WD}
    \vspace{-1.7mm}
\end{minipage}\hfil \ \ 
\begin{minipage}{0.43\linewidth}
   %
    \centering
    \includegraphics[clip, trim=0cm 6.5cm 0cm 7.5cm, width=0.45\textwidth]{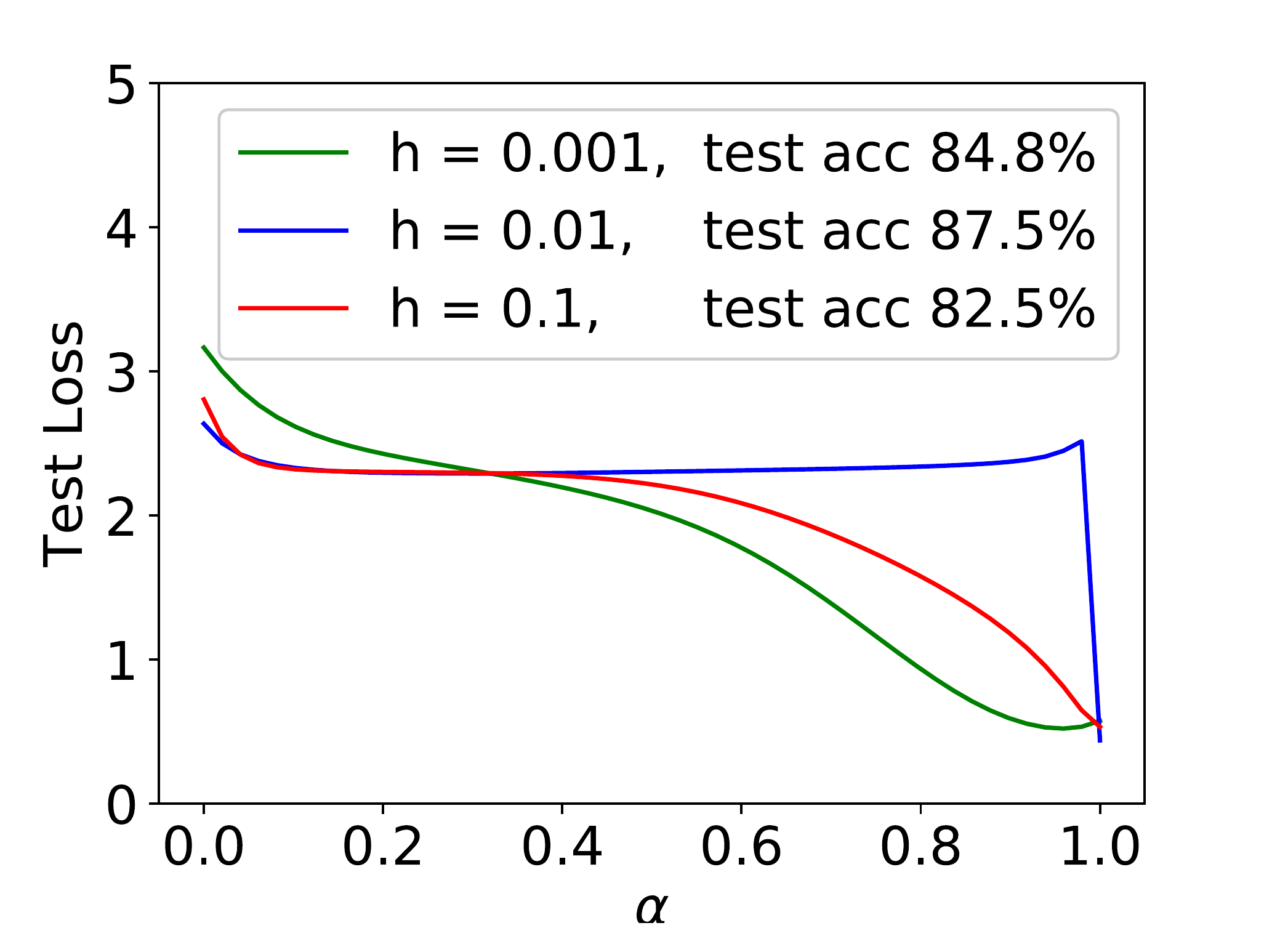}%
   \vspace{-5mm}
    \caption{ResNet-18 trained on CIFAR-10 with fixed learning rate $h$. Averaged over 10 runs.
    }
    \label{fig:fixedlr}
     \centering   
    \includegraphics[clip, trim=0cm 0.35cm 0.2cm 0.2cm, width=0.97\linewidth]{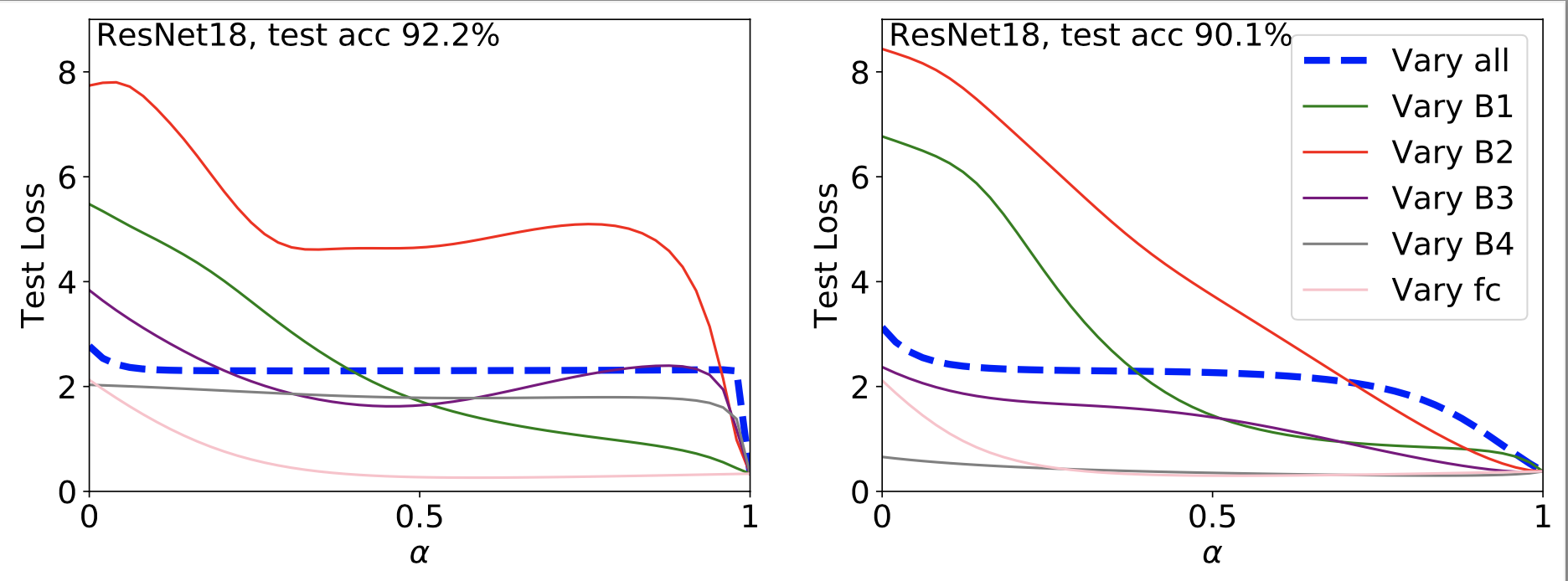} 
    \vspace{-4.5mm}
    \caption{Block-wise linear interpolation for ResNet-18 
    trained 
    on CIFAR-10 using different initial learning rates $h_0$. 
    Left: $h_0 = 0.1$ for all, right: $h_0$ = 0.01 for layers/conv. blocks that exhibit barriers 
    and $h_0$ = 0.1 for the rest. 
    }
    \label{fig:linearintp_difflr}
  \end{minipage} 
  
\vspace*{-0.42cm}
\end{figure*}

\section{The Role of the Optimizer}\label{sec:optimizer}
Optimizer hyperparameter settings, such as the learning rate and weight decay values, strongly affect which optimum is found and the test accuracy. Further, \citet{Lucas2021} found that training modern vision networks with Adam (as opposed to the more typical SGD with momentum) more frequently leads to non-monotonic decay of the loss when linearly interpolating.
Adam also 
increases the distance that the network travels from initialization, leading \citet{Lucas2021} to posit
``large distances moved in weight space encourage non-monotonic interpolation''. 
Inspired by this, we study how using different levels of weight decay or different fixed learning rates ---both of which affect the 
distance travelled and final accuracy--- affect the shape of loss when linearly interpolating.
We find that the hypothesis of \citet{Lucas2021} does not hold in general and that the level of weight decay directly controls the behaviour along the linear path.

\textbf{The effect of weight decay on linear interpolation.}
In Figure \ref{fig:WD}, we vary the amount of weight decay (WD) that is used to train ResNet-18 on CIFAR-10 data.
When using little or no weight decay, the behaviour of the loss varies over different runs; it sometimes exhibits barriers and sometimes monotonic decay.
When using more weight decay, the behaviour changes. At 1e-4 (purple), the loss consistently monotonically decays.
The highest test accuracy is reached when training with 5e-4 (blue), for which loss consists of a plateau with a sudden drop close to the final state. When increasing beyond that, test accuracy decreases, but the loss still exhibits a plateau. The shape of the linear path is thus not a reliable indicator of test accuracy of the final model.
  
\textbf{How do barriers connect to the distance travelled?}
Non-monotonic behaviour occurs when training a model both with zero and large weight decay values (Figure \ref{fig:WD}), while we would expect models without weight decay to travel farther. To further study this relationship, we trained a ResNet-18 at different fixed learning rates $h$ (Figure \ref{fig:fixedlr}).
The linear paths for models trained using higher and lower learning rates ($h = 0.1$ and $0.001$) exhibit monotonic decay, while training with an intermediate learning rate ($h = 0.01$) does not.
This contradicts the hypothesis of \citet{Lucas2021}, under which we would expect the model with the highest learning rate (which travels the furthest) to have a barrier, not a model with a lower learning rate.\footnote{The measure of distance is detailed in Appendix \ref{sec:appxopt}.}
In addition, these results also cast doubt on a connection between monotonically decreasing loss and better accuracy; the middle learning rate (the one that induces a barrier) reaches higher test accuracy than either of the other learning rates (which do not).

\textbf{Layer-wise sensitivity to learning rate.} 
We have observed that linear interpolation behaviour varies by layer (Figure \ref{fig:mnistspiralresnet}B) and that layers/individual conv blocks have different levels of sensitivity to the choice of initialization (Figure \ref{fig:pretrainsomelayers} and Figure \ref{fig:advinitsomelayers}).
This raises the question of how the use of different optimizer hyperparameter choices for different layers affects the shape of the linear path and the final test accuracy of the model. 
Figure \ref{fig:linearintp_difflr} shows the effect of training layers that exhibit barriers with a different learning rate than those that do not. Lowering the initial learning rate used for layers that exhibit barriers removes those barriers, but also substantially lowers the test accuracy of the final model.\footnote{In Appendix \ref{sec:appxopt} we show that lowering the initial learning rate or eliminating weight decay for layers without barriers has a small (ResNet-18) or no (VGG-11) effect on the test accuracy of the trained model, 
whereas doing so for layers with barriers or for all layers lowers test accuracy substantially.} 

\textbf{Summary.} The distance between initial and final state is not a reliable indicator of non-monotonic behaviour. The shape of the linear path is directly influenced by the amount of weight decay and is not indicative of test accuracy. 
Training layers that exhibit barriers with a smaller initial learning rate removes the barriers, but also lowers test accuracy.

\section{The Role of the Model}\label{sec:model}\vspace{-1mm}

Throughout this work we focused on a ResNet-18 architecture with batch normalization. We now discuss the role of the model architecture on the shape of the linear path. 

\begin{figure*}[t]
    \centering
    \includegraphics[width=0.6\linewidth]{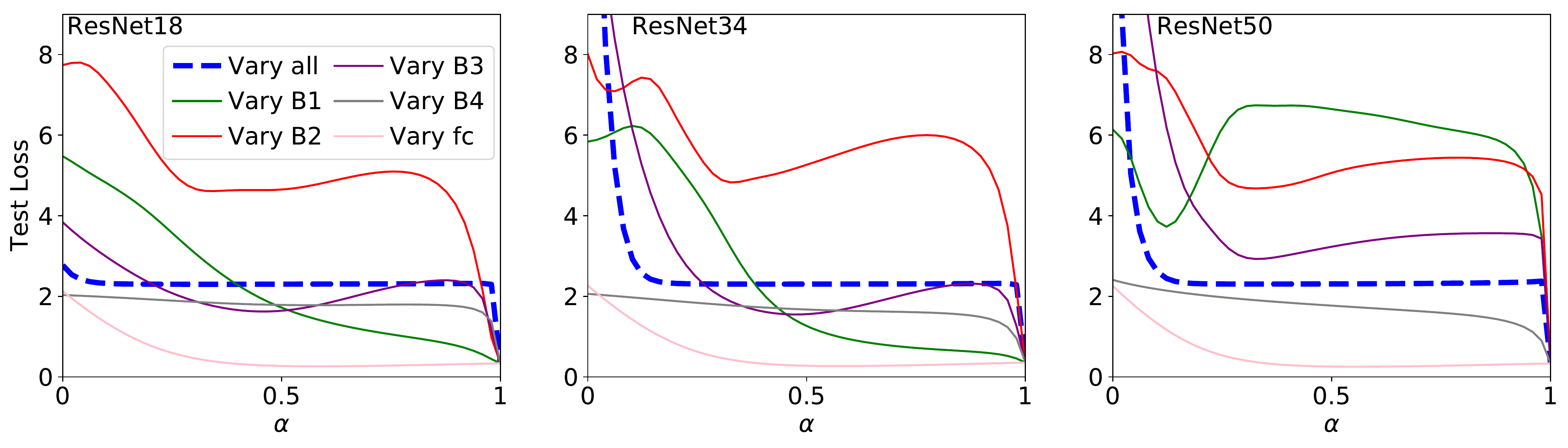}
                \vspace*{-6mm}
    \caption{Test loss when interpolating for convolutional blocks (denoted as B) and the entire network (blue dashed line) for 
    different ResNet architectures trained from scratch on CIFAR-10 data. Results are averaged over 10 runs.}
    \label{fig:compareresnets}
    \vspace{-4mm}
\end{figure*}

\textbf{Architecture depth.}
Generally, the use of deeper ResNet architectures on CIFAR-10 data generates 
or increases the height of convolutional block-wise barriers (Figure \ref{fig:compareresnets}).

\textbf{Role of batch normalization.} Similar to \citet{Lucas2021}, we find that the use of batch normalization (BN) leads to non-monotonic behaviour along the linear path 
for ResNet and VGG architectures (see Appendix \ref{sec:appxmodel}). 
However, we find that for various datasets, MLPs do exhibit barriers in the absence of BN (see e.g., Figure \ref{fig:mnistspiralresnet}A). We also study the linear interpolation parameter group-wise and show that the running mean and variance BN parameters play an important role in the shape of the linear interpolation path observed for the full convolutional block (see Appendix \ref{sec:appxmodel}).

\textbf{Summary of other architectures.} 
In Appendix \ref{sec:appxmodel}, we show that the behaviour for VGG-11 is similar to that of a ResNet-18. For MLPs, we find that increasing the number of nodes in layers with barriers removes the overall presence of barriers.
This effect does not transfer to ResNet architectures. 
For MLPs, early layers contain barriers, whereas later layers do not exhibit barriers. 
Similarly, \citet{Zhang2019} found that the first layer is most sensitive to re-initialization. 

\textbf{Summary.} 
Batch normalization encourages non-monotonic behaviour.
Meanwhile, in Figure \ref{fig:linearintp_difflr} (and Table A\ref{table:difflr}) 
we showed that using a large initial learning rate for layers/conv blocks that contain barriers is necessary to achieve good 
test accuracy. 
These effects together corroborate the observation by \citet{Jastrzebski2020} that using a large learning rate is necessary for networks with batch normalization. We also find that the behaviour for MLPs is distinct from ResNet or VGG architectures. 

\section{Discussion and Future Work}\label{sec:discussion}\vspace{-1mm}

Throughout this work, we explored the relationship between the shape of the linear path between the initial and final model states and the outcome of optimization.
We also studied the linear path in a layer-wise fashion to illustrate that individual convolutional blocks have different sensitivities to initialization and optimizer hyperparameter settings.

\textbf{What influences the shape of loss when linearly interpolating?}
This one-dimensional slice of the landscape is heavily influenced by both initialization and optimization choices.
We found that the data also directly influences the linear path (Figure \ref{fig:mnistspiralresnet}A and Section \ref{sec:data}). 
In addition, the natural language processing literature suggests that attention-based models need adaptive optimizers (e.g. Adam) to train properly \cite{BERT,roberta,AdamvsSGD}, unlike convolutional models for vision data, which can be trained well with SGD with momentum \cite{googlenet,resnet,wideresnet}. While we focused on the latter, we think that revisiting this study for attention-based models is an interesting direction for future work. 

\textbf{Does the shape of loss when interpolating provide insight into other aspects of optimization?}
Linear interpolation is a one-dimensional view of the high-dimensional optimization landscape.
The network follows a different, nonlinear path from initialization to its final weights.
As such, linear interpolation inherently sacrifices information about the optimization problem; prior work and this paper consider whether the information it gleans still provides useful insights into optimization.

Despite the intuition provided by \citet{Goodfellow} that the absence or presence of obstacles reflects difficulty of optimization, we found many cases where higher layer-wise barriers or the creation of barriers along the linear path of the full model appear alongside increased test accuracy.
This evidence implies that either a more difficult optimization trajectory can lead to improved generalization, or Goodfellow et al.'s intuition about the relationship between interpolation and optimization difficulty does not apply to the networks we studied.
Moreover, we found cases where monotonically decreasing loss was accompanied by higher test accuracy (e.g., pre-training).
We also found others where  the linear path exhibited both barriers and monotonic decay over different runs (e.g., weight decay) and where the presence or absence of barriers was not correlated with increased or decreased test accuracy.
In short, for the settings we considered, the shape of the linear path between initial and final state was not a reliable indicator of test accuracy of the trained model.
Our results show that caution is needed when using linear interpolation to make broader claims on the shape of the landscape and success of optimization. 

\textbf{Layer-wise interpolation.}
Interpolating layer-wise makes it possible to study changes in behaviour that are difficult or impossible to discern when interpolating using the full network, for example changes in the data (Figure \ref{fig:numberoftrainingdatadataaug}) or model (Figure \ref{fig:compareresnets}). 
Further, independent of the shape of the linear path, we found that partial pre-training of convolutional blocks can counterintuitively lower test accuracy of the final model; this result implicates the increasingly popular direction of training individual layers in different ways \cite{LARS,LMV,murfet2020,LAMB} and raises questions about the basis upon which we should make such decisions.

\section{Conclusion}
We conclude that the shape of the linear path from initial to final state is \textit{not} a reliable indicator of test accuracy. Although focusing on one line of analysis, e.g. on the role of initialization, seems to suggest that there does exist a connection, the full picture (Table \ref{table:overview}, Fig. A\ref{fig:overview}) illustrates that this is misleading. We believe publishing this negative result is important due to the widespread use of the linear interpolation method. Further, we introduce a new line of inquiry by studying the role played by individual layers and substructures of the network. We find that certain layers require larger initial learning rates to maintain the same test accuracy. 
We also show the surprising adversarial effect of partial pre-training. Further exploration of the layer-wise sensitivity to choice of initialization offers an exciting direction for future work.

 \section*{Acknowledgements}
 Tiffany Vlaar is supported by The Maxwell Institute Graduate School in Analysis and its Applications, a Centre for Doctoral Training funded by the UK Engineering and Physical Sciences Research Council (grant EP/L016508/01), the Scottish Funding Council, Heriot-Watt University and the University of Edinburgh.
 
\bibliography{ref}
\bibliographystyle{icml2022}

\newpage
\appendix
\onecolumn
\section{Implementation Details}\label{sec:implementation}
Throughout the paper we focus on a ResNet-18 architecture with batch normalization trained on CIFAR-10 data. We set batchsize to 128, use cross entropy loss, and use as data augmentation: horizontal flip, random crop, and normalization. We train the network for 100 epochs using SGD with momentum with weight decay set to 5e-4 and the momentum hyperparameter set to 0.9. We use an initial learning rate of $h = 0.1$ that is dropped by a factor of 10 every 33 epochs. For pre-trained models we use initial learning rate $h = 0.001$ that is dropped by a factor of 10 after 30 epochs. We obtain our pre-trained models from the PyTorch library, which have been trained on ImageNet. We replace the final fully connected layer to match the number of classes and (when training on CIFAR-10 data) set the first convolutional layer to have 3 input channels, 64 output channels, and a kernel size of $3\times 3$. We perform all our experiments in PyTorch using NVIDIA DGX-1 GPUs and use standard random PyTorch initialization. Our results are all averaged across multiple runs with different random seeds on initialization and data order.

In this work we vary specific aspects of the training problem, such as the optimizer hyperparameters, data, initialization, and model, while keeping all other aspects fixed. For example, we consider deeper ResNet architectures (Figure \ref{fig:compareresnets}), 
but use the same optimizer, data, and initialization settings as for our base model.
But, as described in the paper, there are a few experiments where we change multiple aspects of the training problem: we turn off weight decay for both our random label initialization experiment (Figure \ref{fig:pretrainedRL}\textit{B}) and our height of barrier initialization experiment (Figure \ref{fig:heightofbarrier_withoutwd}). For the former, we wanted to study the effect of adversarial initialization that lowered the test accuracy of the final trained network. A network trained on 100\% random labels as initialization typically does not affect the test accuracy of the final model, when using SGD with momentum, learning rate decay and weight decay \cite{badglobalminima}. But turning off weight decay during training in combination with pre-training on random labels does lower the test accuracy. For our height of barrier initialization experiment (Figure  \ref{fig:heightofbarrier_withoutwd}) we needed a setting for which the linear path exhibited clear barriers. For a ResNet-18 this is found for some seeds when weight decay is turned off (Figure \ref{fig:WD}). For the runs that exhibited barriers we saved the state of the model at the height of this barrier and used this as initialization for our model to produce Figure  \ref{fig:heightofbarrier_withoutwd}. 

Further, to obtain Figure 1\textit{A} we used a two hidden layer perceptron with 50 nodes in each hidden layer and ReLU activations. We used Adam with $h =$ 5e-4, without weight decay or any learning rate scheduling. We use batch size 128 for MNIST data and cross entropy loss. We use binary cross entropy loss for the binary classification spiral data set. 
The first class of the spiral data set is generated using
\begin{align}
 x& = 2 \sqrt{t} \cos(8 \sqrt{t} \pi) +0.02\mathcal{N}(0,1), \nonumber \\
 y& = 2 \sqrt{t} \sin(8 \sqrt{t} \pi) +0.02\mathcal{N}(0,1), \label{spiraleqn}
\end{align}
where $t$ is drawn repeatedly from the uniform distribution $\mathcal{U}(0,1)$ to generate data points. The other class of this dataset is obtained by shifting the argument of the trigonometric functions by $\pi$. For our experiments on the spiral data set we used 10000 training data, 5000 test data points and a batch size of 500.  

\makeatletter
\renewcommand{\fnum@figure}{\figurename~A\thefigure}
\makeatother

\section{Overview of Results}\label{sec:overview}
\begin{figure}[h]
    \centering
    \includegraphics[width=0.8\linewidth]{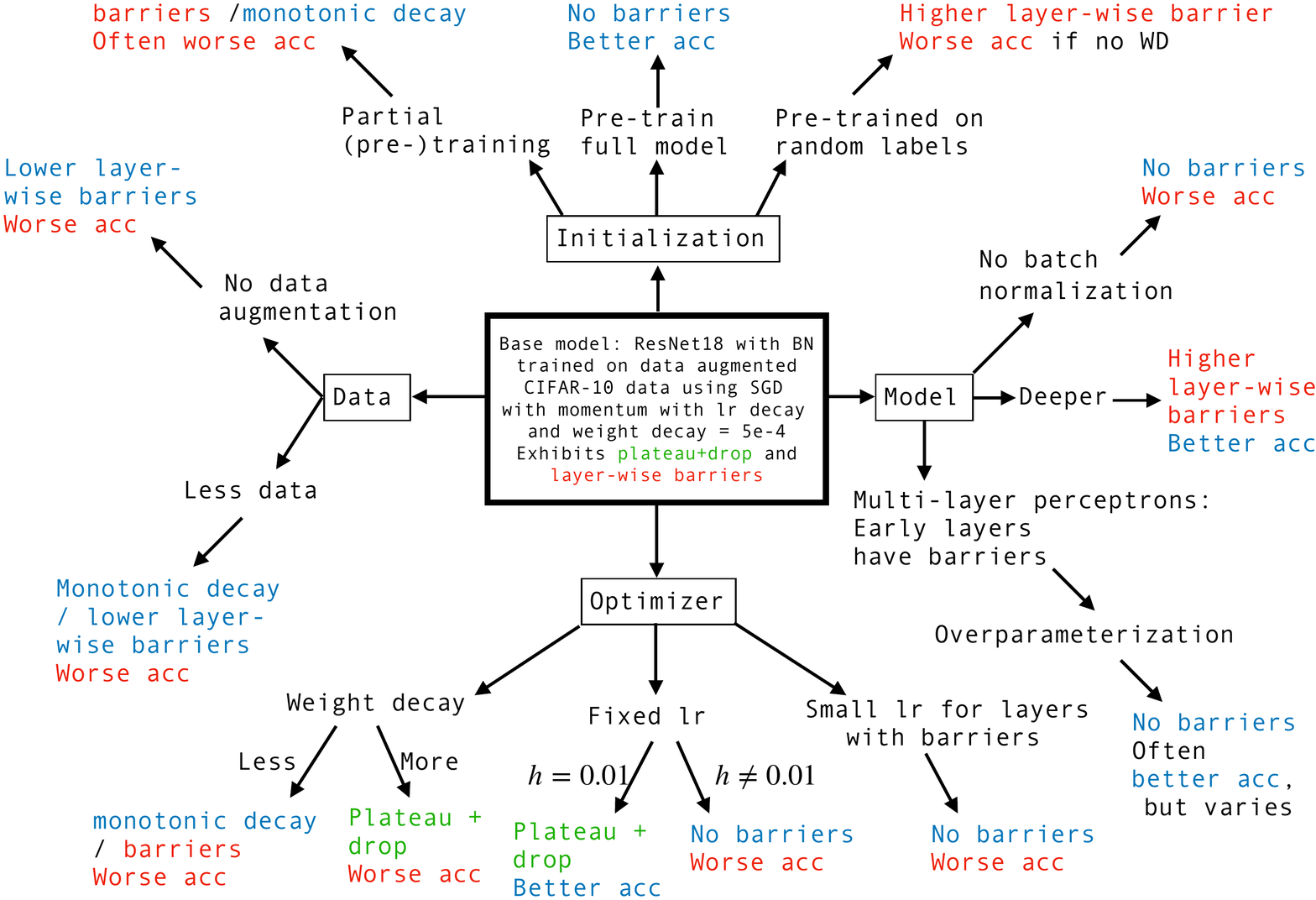}
    \caption{The linear path between $\theta_i$ and $\theta_f$ of our base ResNet-18 architecture exhibits a plateau at the level of random chance until close to the optimum (plateau + drop); it exhibits barriers when evaluating the linear path layer-wise (see Figure 1\textit{B}). The base model obtains 92.2\% accuracy on CIFAR-10 test data. In this figure we present the change in test accuracy and shape of the linear path between $\theta_i$ and $\theta_f$ compared to this base model along four axes: the optimizer, the data, the initialization, and the model. The presence of barriers (indicated in red) frequently coincides with a better test accuracy (indicated in blue). }
    \label{fig:overview}
\end{figure}

A summary of all our findings can be found in Figure A\ref{fig:overview}. 

\newpage 
\section{Further Studies on the Role of Initialization}\label{sec:appxinit}
\begin{figure}[h]
    \centering
    \includegraphics[width=\linewidth]{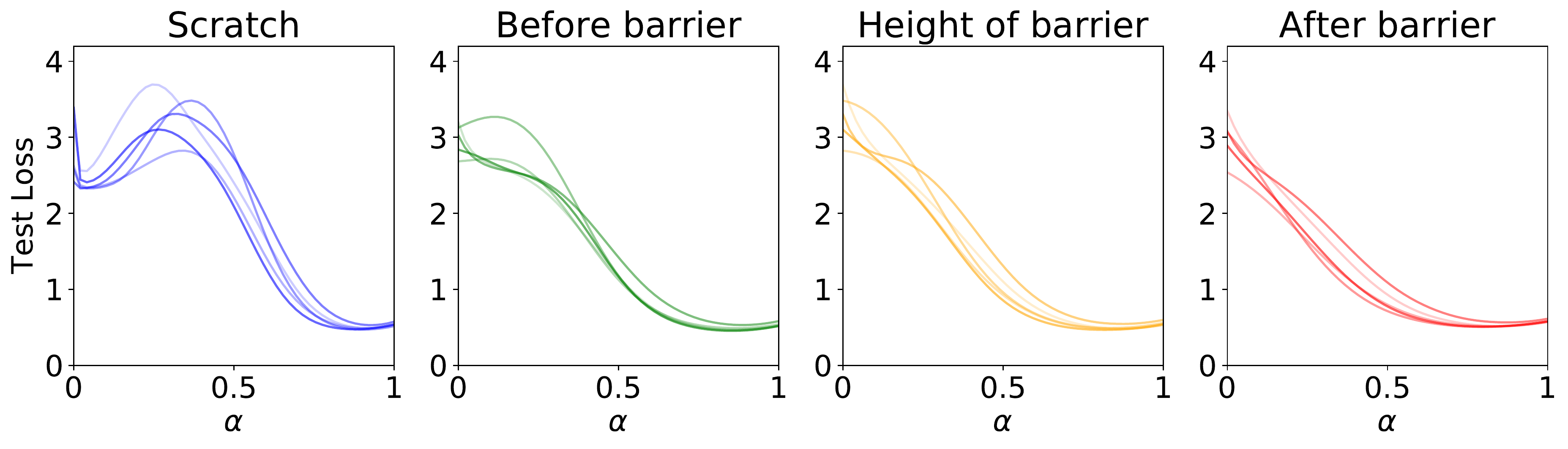}
    \caption{ ResNet-18 on CIFAR-10 without weight decay. Test loss between $\theta_i$ and $\theta_f$ for discrete set of $\alpha\in [0,1]$ in 50 equally spaced subintervals, when trained either from scratch (blue) or from an initialization corresponding to $\theta_{\alpha}$ of the blue line at: the height of the barrier $\alpha_{\text{barrier}}$ (orange), just before the barrier at $\alpha = \alpha_{\text{barrier}}-5$ (green), just after the barrier at $\alpha = \alpha_{\text{barrier}}+5$ (red).}
    \label{fig:leftrightofbarrier}
\end{figure}

\textbf{Left/right of barrier initialization.}
In Section \ref{sec:initialization} we showed the effect of initializing neural networks at the
parameter configuration which corresponds to the height of the barrier along their initial to converged state. We considered a ResNet-18 architecture which was trained without weight decay on CIFAR-10 data. The linear interpolation path in this setting clearly exhibits barriers for some seeds (Section \ref{sec:optimizer}, Figure \ref{fig:WD}). For the runs that exhibited barriers, we saved the model state at the the height of the barrier and used this as initialization for our model.  Although it was initialized at a higher loss than occurs at random initialization, the resulting network obtained a higher test accuracy than the network trained from scratch and its linear path exhibited monotonic decay (Figure  \ref{fig:heightofbarrier_withoutwd}). In 
Figure A\ref{fig:leftrightofbarrier} and Table A\ref{table:leftrightofbarrier} we show that these effects are not restricted to initialization at the barrier: choosing an initialization along the linear path to the left or the right of the barrier leads to similar performance improvement (Table A\ref{table:leftrightofbarrier}) and typically monotonic decay along the linear path (Figure A\ref{fig:leftrightofbarrier}).

\makeatletter
\renewcommand{\fnum@table}{\tablename~A\thetable}
\makeatother

\begin{table}[h]
\centering
\caption{ ResNet-18 on CIFAR-10 without weight decay. Test loss between $\theta_i$ and $\theta_f$ for discrete set of $\alpha\in [0,1]$ in 50 equally spaced subintervals, when trained either from scratch (blue) or from an initialization corresponding to $\theta_{\alpha}$ of the blue line at: the height of the barrier $\alpha_{\text{barrier}}$, before the barrier at $\alpha = \alpha_{\text{barrier}}-5$, after the barrier at $\alpha = \alpha_{\text{barrier}}+5$ (red). \\}\label{table:leftrightofbarrier}
\begin{tabular}{c|c|c}
  Initialization & Training accuracy (\%) & Test accuracy (\%)  \\ \hline\hline
  Random initialization   &  99.44 $\pm$0.12 & 90.01 $\pm$0.39 \\ \hline
  Before barrier & 99.80 $\pm$0.04 & 91.05 $\pm$0.21\\
  Height of barrier & 99.86 $\pm$0.02 & 91.12 $\pm$0.18\\
  After barrier & 99.88 $\pm$ 0.02 & 90.85 $\pm$0.17 \\
\end{tabular}
\end{table}

\begin{figure}[t]
    \centering
    \includegraphics[width=\linewidth]
    {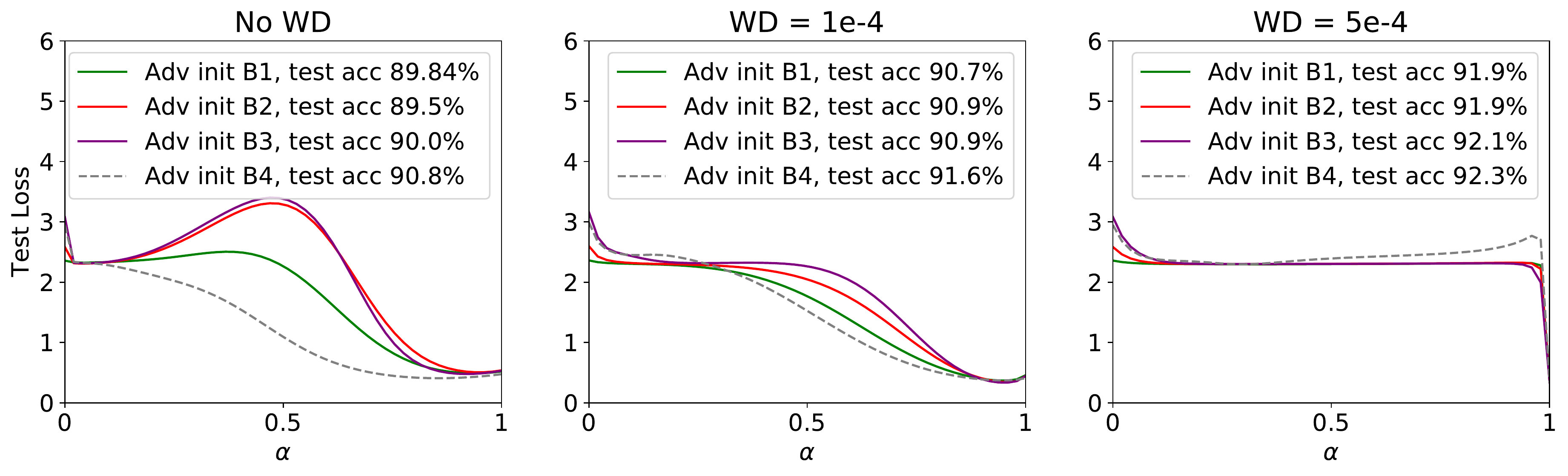}
   \vspace{-6mm}
    \caption{ Test loss between $\theta_i$ and $\theta_f$ 
    when one specific conv block (B) was adversarially initialized using random labels. The net was trained either without weight decay (WD) (left), with WD = 1e-4 (middle) or with WD = 5e-4 (right). Averaged over 10 runs.}
    \label{fig:advinitsomelayers_varyWD} 
\end{figure}

\textbf{Block-wise adversarial initialization with different levels of weight decay.} 
To further explore the effect of using different initializations for individual convolutional blocks on the linear path and test accuracy, we set one convolutional block to a random label adversarial initialization while using a standard random initialization for the rest of the net. We then re-train using this initialization with different amounts of weight decay (WD). We vary the WD value to study how changes in the full model linear path (Figure \ref{fig:WD}) are reflected by the convolutional block-wise linear path. 
We find that while adversarial initialization for the first convolutional blocks lowers the test accuracy, using adversarial initialization for convolutional block 4 slightly increases the final test accuracy compared to training from scratch (compare Figure A\ref{fig:advinitsomelayers_varyWD} with Figure \ref{fig:WD}). 
This is also reflected by the shape of the linear path: when using an adversarial initialization for convolutional block 4 the linear path exhibits monotonic decay while others settings exhibit barriers (no WD) and a barrier while other settings exhibit a plateau (WD = 5e-4).

\section{Further Studies on the Role of Data}\label{sec:appxdata}

\begin{figure}[h]
    \centering
        \includegraphics[width=\linewidth]{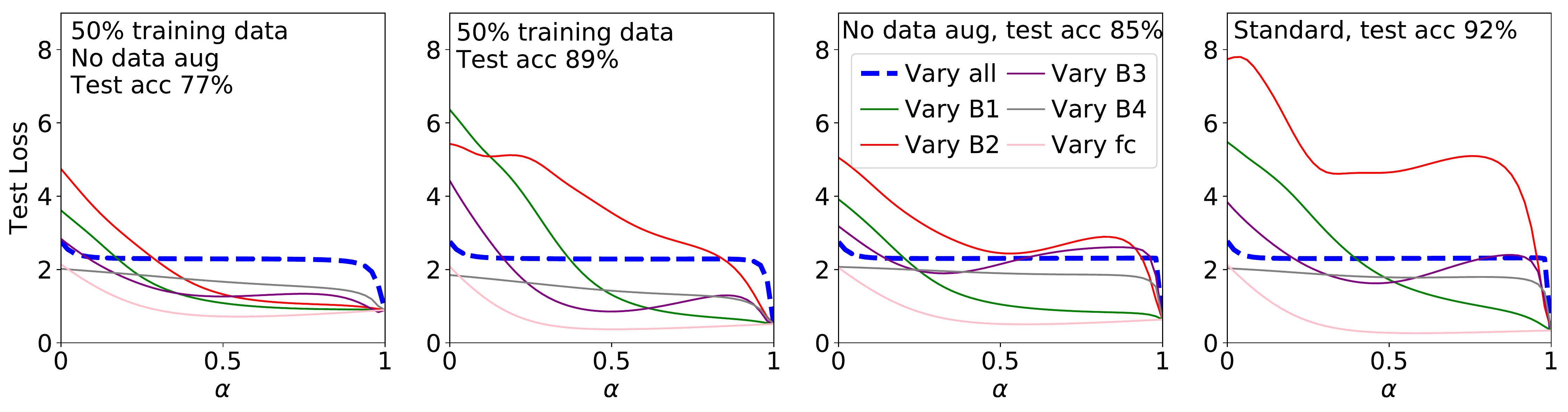}
    \caption{Test loss between $\theta_i$ and $\theta_f$ for ResNet-18 on CIFAR-10 when varying the amount of training data and whether data augmentation (horizontal flips and random crops) is used.}
    \label{fig:lessdataandnoaug}
\end{figure}

\textbf{Less training data and no data augmentation.} 
Further to our experiments in Section \ref{sec:data}, we here show the combined effect of reducing the number of training data in addition to removing data augmentation (Figure A\ref{fig:lessdataandnoaug}, left). We find that although this further reduces the test accuracy and presence of block-wise barriers, the overall behaviour for the linear interpolation of the full model (blue dotted line) is maintained (unless the number of training data is further reduced, see Figure \ref{fig:numberoftrainingdatadataaug}, left in main paper).

\begin{figure}[h]
    \centering
        \includegraphics[width=\linewidth]{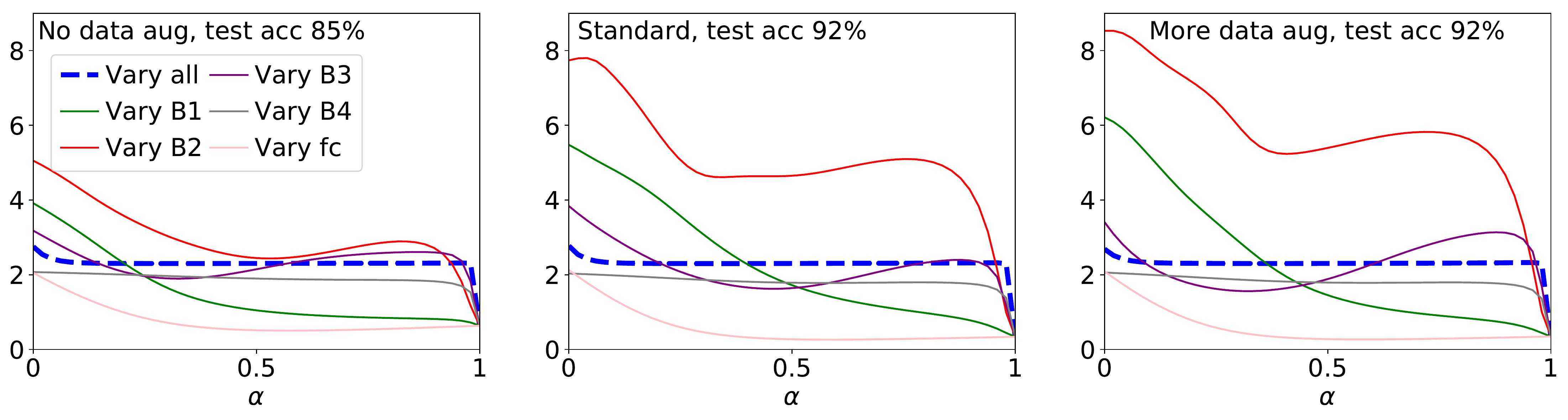}
    \caption{Test loss between $\theta_i$ and $\theta_f$ for ResNet-18 on CIFAR-10 when using either no data augmentation (left), standard data augmentation (horizontal flips and random crops) (middle), or additional data augmentation (brightness = 0.05) (right).}
    \label{fig:lessandmoreaug}
\end{figure}

\textbf{More data augmentation.} In Figure A\ref{fig:lessandmoreaug} we show that using additional data augmentation causes the height of layer-wise barriers to slightly increase (for B2 and B3). 

\newpage
\section{Further Studies on the Role of Optimization} \label{sec:appxopt}
\textbf{Distance.} In our fixed learning rate experiment (Figure \ref{fig:fixedlr}) we showed that setting $h = 0.01$ leads to the largest test accuracy of the trained model and non-monotonic behaviour of the loss along the linear path between initial and final state. Meanwhile using a larger ($h = 0.1$) or smaller ($h = 0.001$) learning rate led to a lower test accuracy and monotonic decay of the loss along the linear path. We used the same initialization to compare the different settings, but averaged the results presented in Figure \ref{fig:fixedlr} over 10 runs using different random seeds. To measure the distance between initial $\theta_i$ and final $\theta_f$ models we use the following measure (as used in \cite{badglobalminima}):
\begin{align}
    d(\theta_f,\theta_i) = \frac{\|\theta_f-\theta_i\|_F}{\|\theta_i\|_F},\label{eq:distance}
\end{align}
where $\|\cdot\|_F$ denotes the Frobenius norm. There exist many other ways of measuring the distance, which affect the order of magnitude. Since we focus only on the relative distances travelled for different learning rate settings the use of Eq. \eqref{eq:distance} suffices for this purpose. We find that the average distance travelled when using $h = 0.1$ is 1.03, whereas the average distance travelled when using $h = 0.01$ is 0.88 and when using $h = 0.001$ is 0.22. This confirms our intuition that using a larger learning rate will increase the distance travelled.

\begin{figure}[h]
    \centering
    \includegraphics[width=0.7\linewidth]{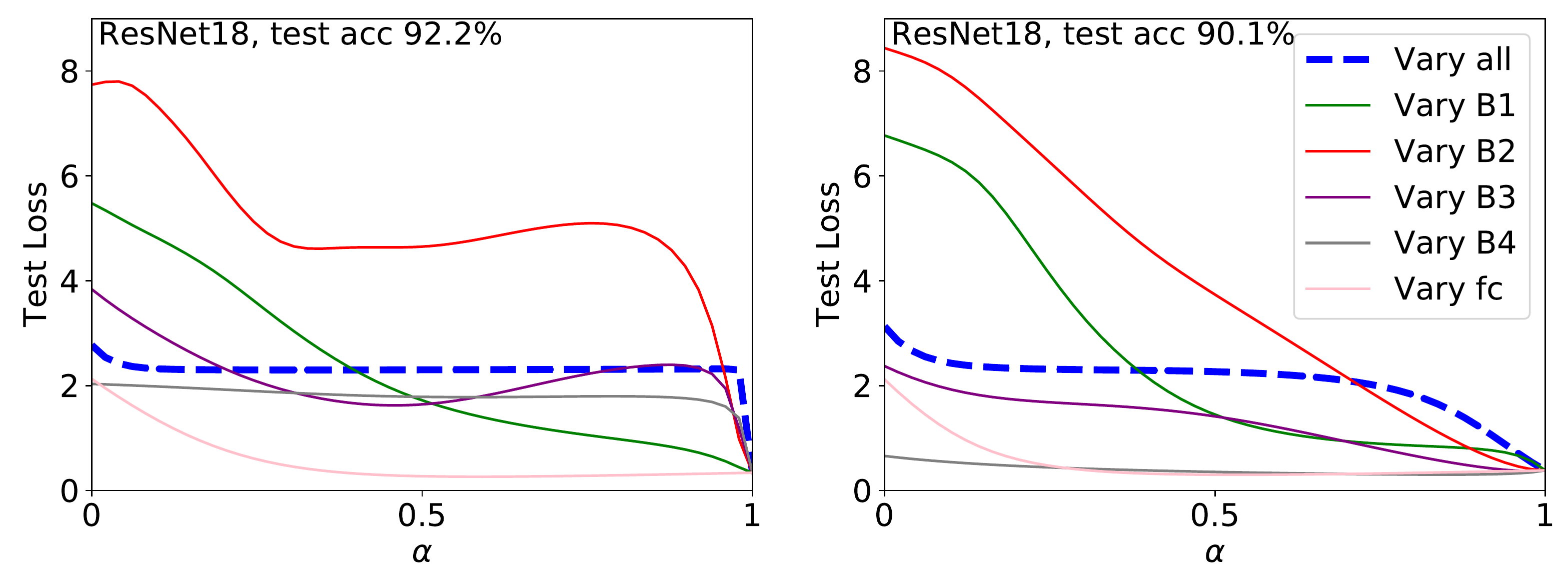}\\
      \includegraphics[width=0.7\linewidth]{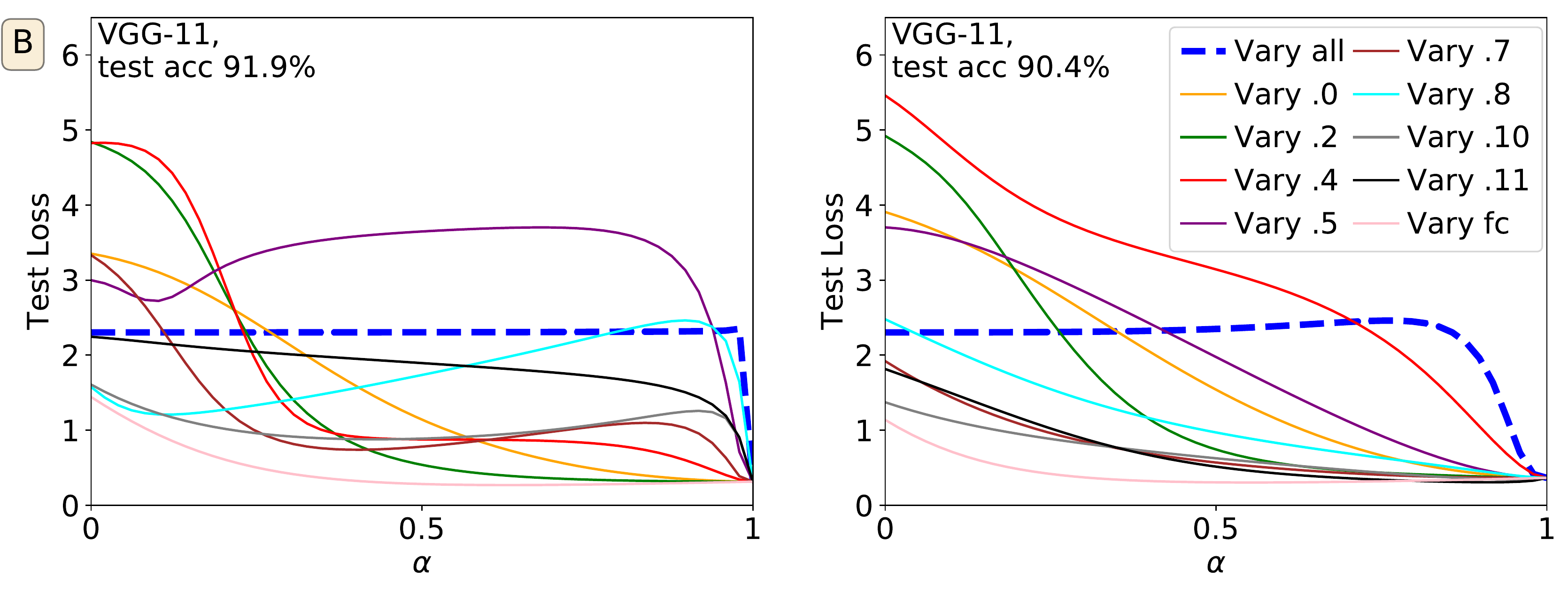}
    \caption{Layer-wise (or block-wise) linear interpolation for ResNet-18 \textit{(A)} and VGG-11 \textit{(B)} trained from scratch on CIFAR-10 data using different initial learning rates $h_0$. For ResNet-18 \textit{(A)}: left: $h_0 = 0.1$ for all, right: $h_0$ = 0.01 for conv block 2, 3, and 4.0 and $h_0$ = 0.1 for rest. For VGG-11 \textit{(B)}: left: $h_0 = 0.1$ for all, right: $h_0$ = 0.01 for layers 4-10 and $h_0$ = 0.1 for rest. Using smaller learning rates for layers that exhibit barriers (ResNet-18: conv block 2, 3, and 4.0, VGG-11: layer 4-10) can remove barriers, but lowers the test accuracy of the resulting model. 
    }
     \label{fig:linearintp_difflrvgg}
\end{figure}

\textbf{Custom optimization schemes.}
We observed that using a 
smaller initial learning rate for layers or convolutional blocks which exhibit barriers removes these barriers, but also lowers the test accuracy of the resulting model for a ResNet-18 architecture (Figure \ref{fig:linearintp_difflr} or Figure A\ref{fig:linearintp_difflrvgg}\textit{A}). 
We observe the same behaviour for a VGG-11 architecture (Figure A\ref{fig:linearintp_difflrvgg}\textit{B}). We consider layers (or convolutional blocks) to not exhibit barriers if the loss along their linear path is monotonically decreasing; the exact layers are specified in the caption of Figure A\ref{fig:linearintp_difflrvgg}. 

The changes in linear interpolation behaviour per layer, raises the question whether we can optimize different layers differently depending on the absence or presence of barriers.
Table A\ref{table:difflr} shows the result of using custom optimization schemes that train layers (or convolutional blocks) that exhibit barriers with a different learning rate or amount of weight decay than those that do not. Lowering the initial learning rate used for layers without barriers or eliminating weight decay has a smaller (ResNet-18) or no (VGG-11) effect on the test accuracy of the trained model, whereas doing so for layers with barriers or for the full model does significantly affect performance.

\begin{table}[h]
    \caption{We train layers that exhibit barriers (\textit{B}) with a different learning rate or amount of weight decay (WD) than those that do not (\textit{NB}). Test accuracy is maintained much better when training only the layers with barriers at the higher learning rate or with weight decay. Results are averaged over 10 runs.}
     \label{table:difflr}
    \centering
    \begin{tabular}{l|c|c}
    & \multicolumn{2}{c}{Test accuracy} \\
    Initial learning rate $h_0$ & ResNet-18 & VGG-11 \\ \hline
    $0.1$ (all) & 92.2 $\pm$0.2\% & 91.9 $\pm$0.2\% \\ \hline
    $0.1$ (\textit{B}), $0.01$ (\textit{NB}) & 91.8 $\pm$0.2\% & 92.0 $\pm$0.1\%\\
    $0.01$ (\textit{B}), $0.1$ (\textit{NB}) & 90.1 $\pm$0.2\% & 90.4 $\pm$0.1\%\\
    $0.01$ (all) & 90.3 $\pm$0.3\% & 90.9 $\pm$0.1\% \\ \hline
    WD (\textit{B}), No WD (\textit{NB}) & 91.5 $\pm$0.2\%  & 91.8 $\pm$0.1\%  \\
    No WD (\textit{B}), WD (\textit{NB})  & 90.1 $\pm$0.3\%  & 90.2 $\pm$0.1\%  \\
    No WD (all) & 90.1 $\pm$0.4\% & 90.4 $\pm$0.3\%  \\
    \end{tabular}
    \vspace{-7mm}
\end{table}

\newpage
\section{Further Studies on the Role of the Model}\label{sec:appxmodel}

\begin{figure}[h]
    \centering
    \includegraphics[width=0.3\linewidth]{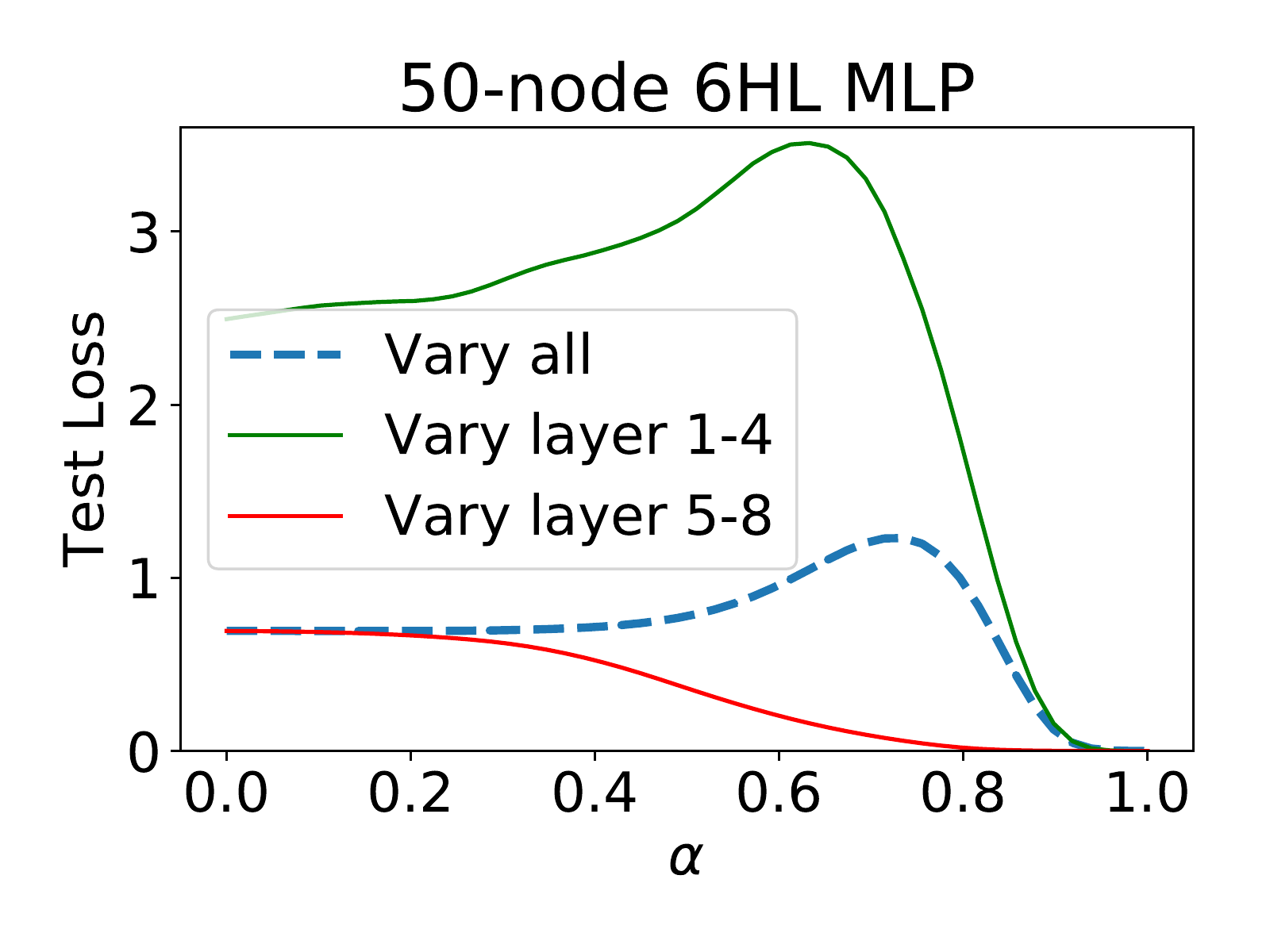}
    \includegraphics[width=0.3\linewidth]{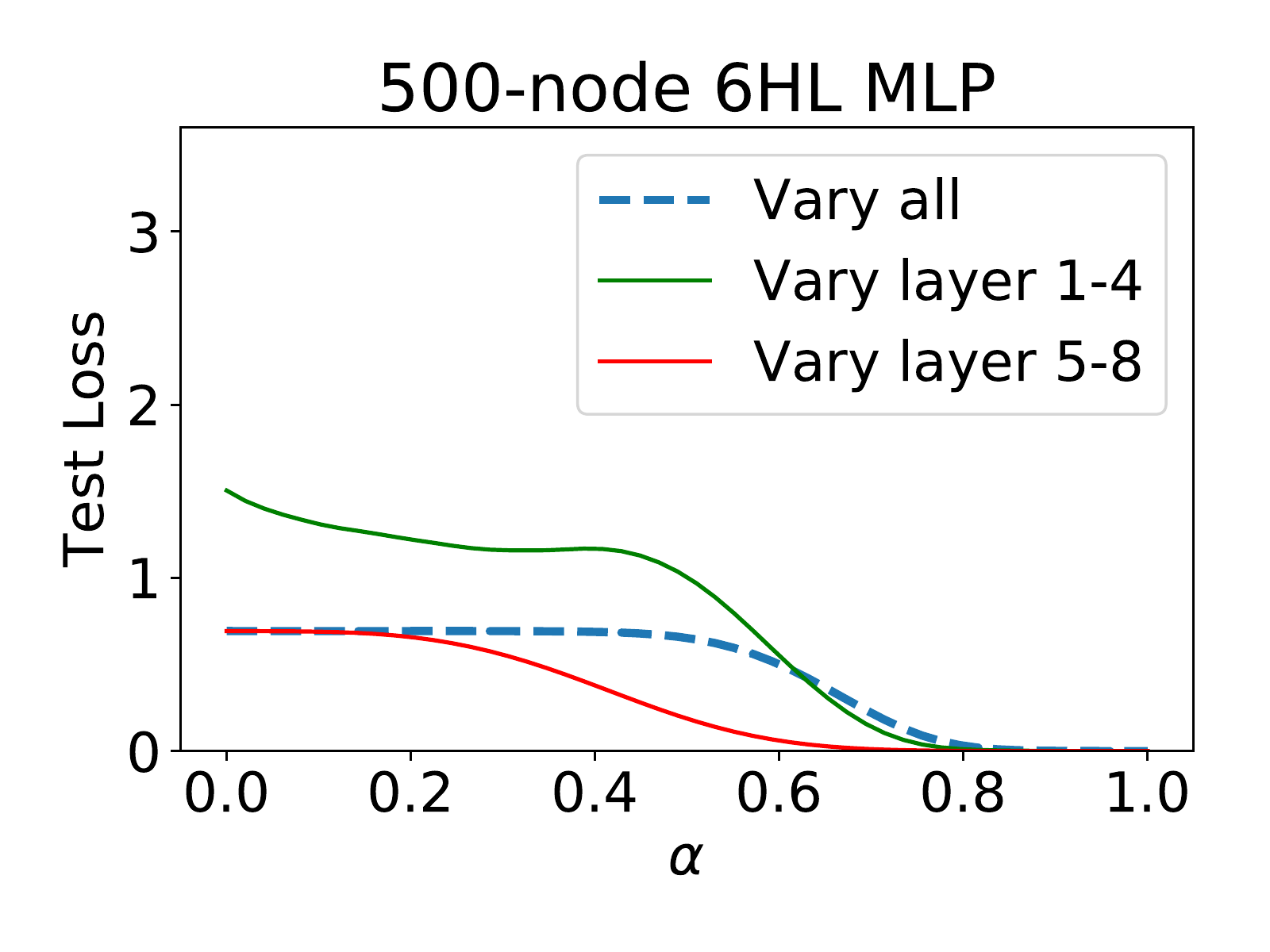}
    \includegraphics[width=0.3\linewidth]{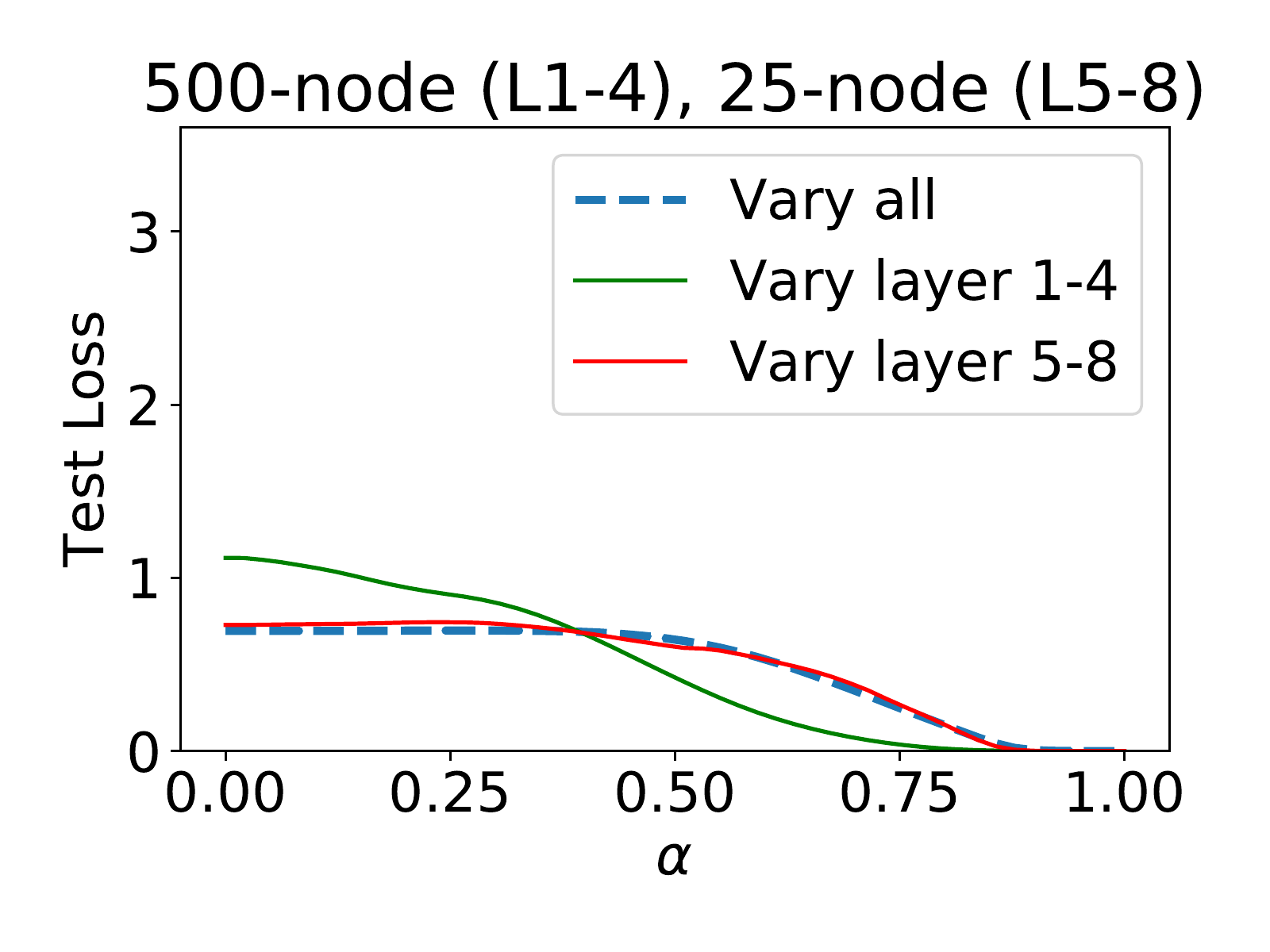}
    \caption{A 6 hidden layer MLP trained on spiral data (generated using Eq. \ref{spiraleqn}) using different amount of nodes in the hidden layers: 50-node for all layers (left), 500-node for all layers (middle), 500-node for layers that exhibit barriers (layer 1-4) and 25-node for layers that exhibit monotonic decay (layer 5-8) (right). All approaches obtain 100\% test accuracy. Increasing the number of nodes in layers that exhibit barriers removes the presence of barriers for the overall model (blue dotted line).}
    \label{fig:mlpspiral}
\end{figure}

\textbf{Multi-layer perceptrons.} For multi-layer perceptrons (MLPs) we find that the early layers (close to the input) typically exhibit barriers, whereas later layers exhibit monotonic decay (Figure A\ref{fig:mlpspiral}, left). By increasing the number of nodes in layers that exhibit barriers one can reduce the presence of barriers for the overall model (Figure A\ref{fig:mlpspiral}, middle). Meanwhile, layers that do not exhibit barriers can be narrowed without removing the monotonic decay property (Figure A\ref{fig:mlpspiral}, right).

\begin{figure}[h]
    \centering
    \includegraphics[clip, trim=0cm 7cm 0cm 8cm, width=0.35\textwidth]{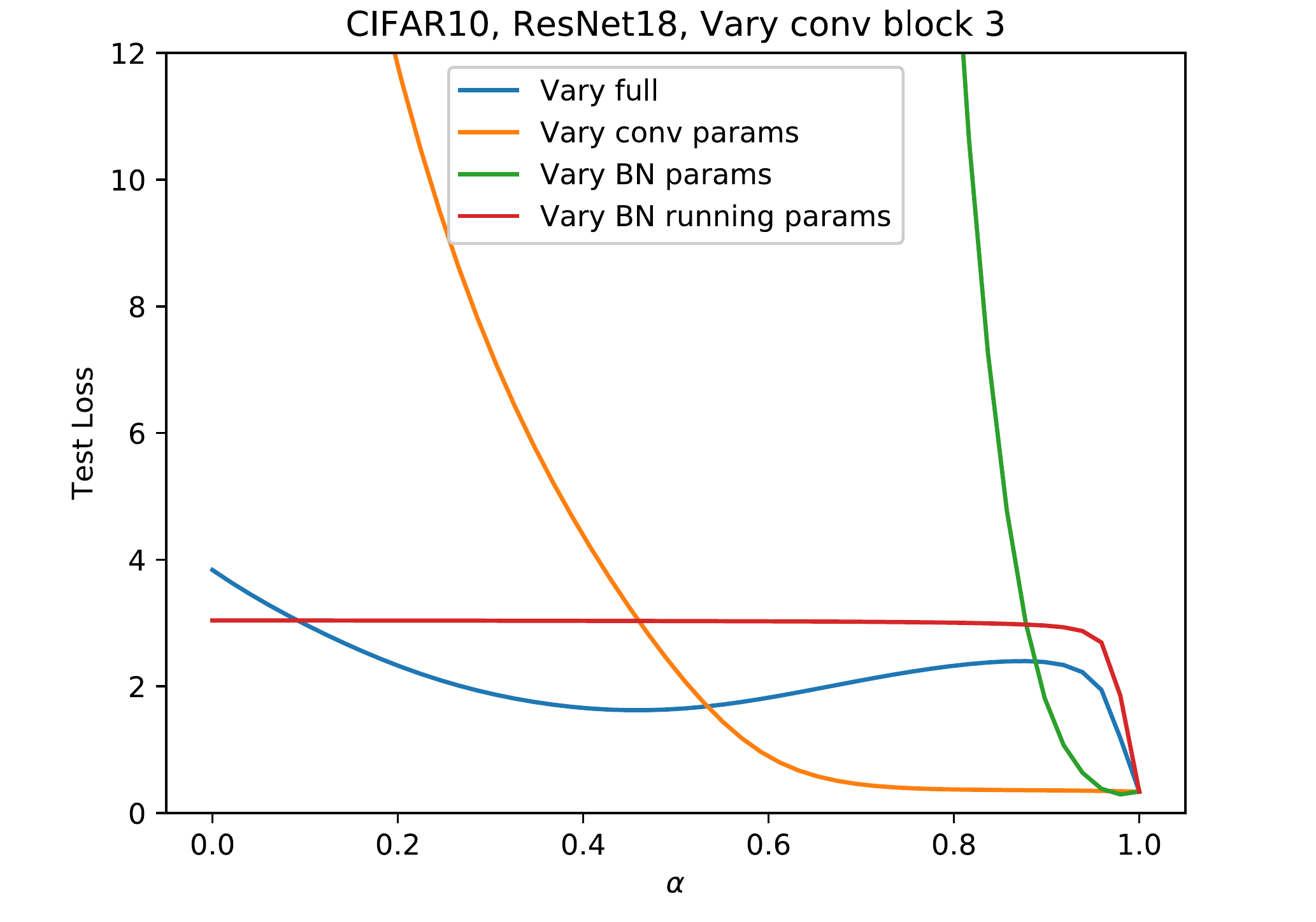}
    \caption{We keep all parameters of a ResNet-18 architecture fixed at their final trained position (on CIFAR-10 data), apart from specific parameter groups in the third convolutional block, which we vary from their initial to final position using the linear interpolation technique. The blue line is when the full third convolutional block is varied. The batch normalization parameters are split into two groups, the running mean and variance (red) and the others (green). }
    \label{fig:roleofBN_convblock3}
\end{figure}

\begin{figure}[h]
    \centering
    \includegraphics[width = 0.7 \linewidth]{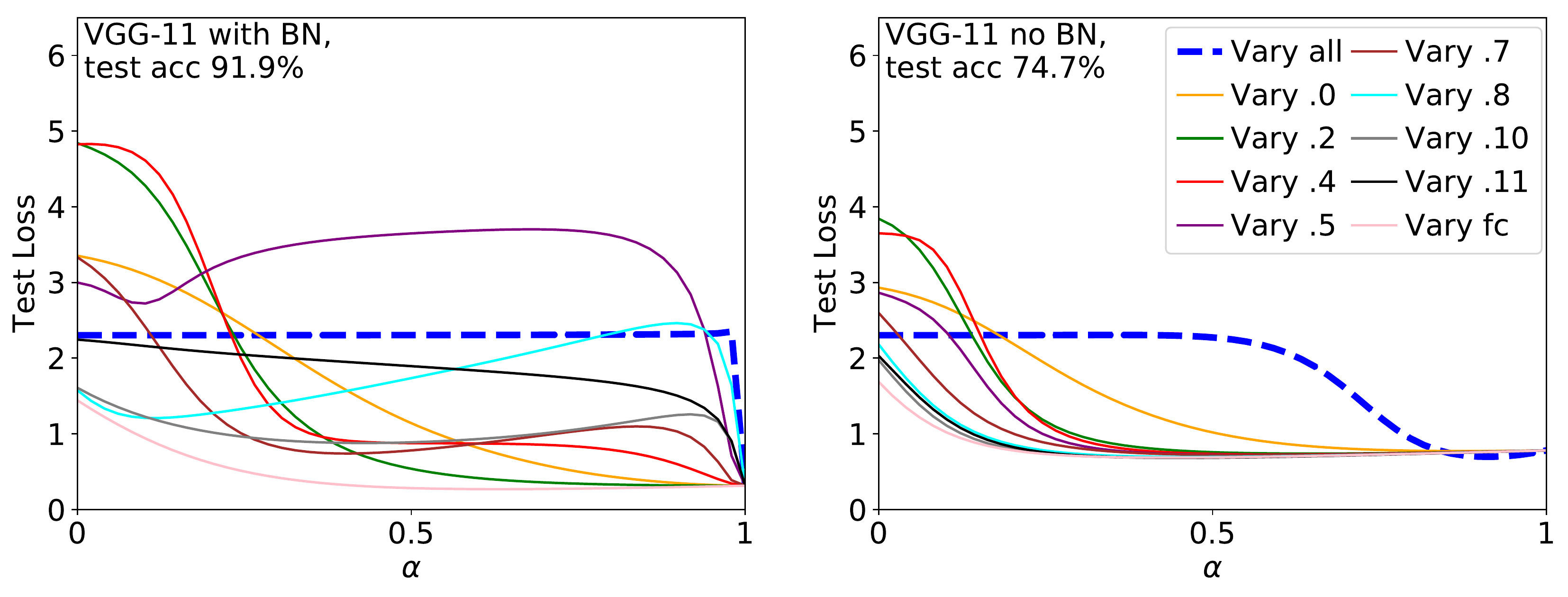}
    \caption{Test loss between $\theta_i$ and $\theta_f$ for VGG-11 on CIFAR-10 when varying whether batch normalization (BN) is used.}
    \label{fig:varyBN}
\end{figure}

\textbf{Batch normalization.}
In Figure A\ref{fig:roleofBN_convblock3} we show using parametergroup-wise linear interpolation that the running mean and variance batch normalization parameters appear to govern the behaviour of the full convolutional block. For the linear path of the full model, we corroborate the findings of \cite{Lucas2021} that the use of batch normalization often leads to non-monotonic behaviour (Figure A\ref{fig:varyBN}). In the absence of batch normalization, ResNet and VGG architectures often exhibit monotonic decay along their linear path from initial to final state. However, for various datasets we find that MLPs do exhibit barriers in the absence of batch normalization, see e.g., Figure A\ref{fig:mlpspiral}.

\textbf{VGG-11.} In Figure A\ref{fig:varyBN} we show the loss along the linear path for a VGG-11 architecture on CIFAR-10 data with (left) and without (right) batch normalization. 

Layer width: We find for a VGG-11 architecture with batch normalization on CIFAR-10 data that layers without barriers can be narrowed without affecting the generalization performance (Table A\ref{tab:narrowVGG11}). This does not hold for layers that do exhibit barriers along the linear path between the initial and final state. We found that this effect did not generalize to a ResNet-18 architecture, so recommend caution with using the presence or absence of layer-wise layers as an indication of which layers can be narrowed. 

\begin{table}[h]
    \centering
        \caption{Layer-wise linear interpolation for a VGG-11 architecture with batch normalization trained from scratch on CIFAR-10 data. This architecture exhibits no barriers in the first two layers, final layer and the fully connected layer, whereas it does exhibit barriers in the five middle layers (see also Figure A\ref{fig:varyBN}, left). We find that narrowing all non-barrier layers by half only marginally lowers the test accuracy (2nd row), whereas narrowing only one barrier layer (3rd row) already has a large impact. Narrowing two (4th row) or all barrier layers (final row) lowers the test accuracy even further.}
    \begin{tabular}{c|c}
    Intervention & Test accuracy (\%) \\ \hline \hline
    Standard & 91.93 $\pm$0.19 \\ \hline
    Narrow no-barrier layers & 91.73 $\pm$0.19\\
    Narrow \ \ 1 barrier layer\  \ & 91.53 $\pm$0.20\\
    Narrow \ \ 2 barrier layers & 91.25 $\pm$0.12 \\
    Narrow all barrier layers & 91.15 $\pm$0.13\\
    \end{tabular}
    \label{tab:narrowVGG11}
\end{table}
\end{document}